\documentclass[10pt,journal,compsoc]{IEEEtran}
\usepackage{epsfig}
\usepackage{graphicx}
\usepackage{amsmath}
\usepackage{amssymb}
\usepackage{multirow}
\usepackage{tabularx, cellspace}
\usepackage[colorlinks,linkcolor=red]{hyperref}

\ifCLASSOPTIONcompsoc
  \usepackage[nocompress]{cite}
\else
  \usepackage{cite}
\fi
\ifCLASSINFOpdf
\else
\fi
\begin{document}
%
\title{Unsupervised Person Re-Identification with Wireless Positioning under Weak Scene Labeling}
%
%
%
%

\author{Yiheng Liu,
Wengang Zhou,
Qiaokang Xie,
Houqiang Li,~\IEEEmembership{Fellow,~IEEE}
\thanks{Yiheng Liu, Wengang Zhou, Qiaokang Xie, and Houqiang Li are with CAS Key Laboratory of GIPAS, University of Science and Technology of China, Hefei, China. Wengang Zhou and Houqiang Li are also with Institute of Artificial Intelligence, Hefei Comprehensive National Science Center. E-mail: \{lyh156, xieqiaok\}@mail.ustc.edu.cn, \{zhwg, lihq\}@ustc.edu.cn.}
\thanks{Corresponding authors: Wengang Zhou and Houqiang Li.}}

%
%

\markboth{Journal of \LaTeX\ Class Files,~Vol.~**, No.~**, **~**}%
{Shell \MakeLowercase{\textit{et al.}}: Bare Demo of IEEEtran.cls for Computer Society Journals}
%



\IEEEtitleabstractindextext{%
\begin{abstract}
  Existing unsupervised person re-identification methods only rely on visual clues to match pedestrians under different cameras. Since visual data is essentially susceptible to occlusion, blur, clothing changes, etc., a promising solution is to introduce heterogeneous data to make up for the defect of visual data. Some works based on full-scene labeling introduce wireless positioning to assist cross-domain person re-identification, but their GPS labeling of entire monitoring scenes is laborious. To this end, we propose to explore unsupervised person re-identification with both visual data and wireless positioning trajectories under weak scene labeling, in which we only need to know the locations of the cameras. Specifically, we propose a novel unsupervised multimodal training framework (UMTF), which models the complementarity of visual data and wireless information. Our UMTF contains a multimodal data association strategy (MMDA) and a multimodal graph neural network (MMGN). MMDA explores potential data associations in unlabeled multimodal data, while MMGN propagates multimodal messages in the video graph based on the adjacency matrix learned from histogram statistics of wireless data. Thanks to the robustness of the wireless data to visual noise and the collaboration of various modules, UMTF is capable of learning a model free of the human label on data. Extensive experimental results conducted on two challenging datasets, i.e., WP-ReID and Campus4K demonstrate the effectiveness of the proposed method.
\end{abstract}

}

\maketitle

\IEEEdisplaynontitleabstractindextext

%
\IEEEpeerreviewmaketitle

\IEEEraisesectionheading{\section{Introduction}\label{sec:introduction}}

%
%
%
%
\IEEEPARstart{P}{erson} re-identification is essentially a person retrieval task in a multi-camera surveillance network. Given an image or video of a person that we are interested in, it aims to find out the images or videos of this person from a large data corpus captured by multiple surveillance cameras. 
The supervised person re-identification \cite{zheng2012reidentification, lisanti2014person, ahmed2015improved, cheng2016person, liu2017end, matsukawa2019hierarchical, wei2020sif, zhang2020person} requires massive and exhaustive identity labeling of the cross-camera data, which is laborious and suffers the scalability issue in real-world applications.
To bypass the label requirement, unsupervised person re-identification \cite{li2018unsupervised, bak2018domain, zhong2018generalizing, deng2018image, liang2018m2m, zhong2019invariance, yu2019unsupervised} is proposed to directly learn models from unlabeled data. Thanks to its favorable potential, it has received substantial attention from both academia and industry in recent years. 

\begin{figure}
   \begin{center}
   \includegraphics[width=0.95\linewidth]{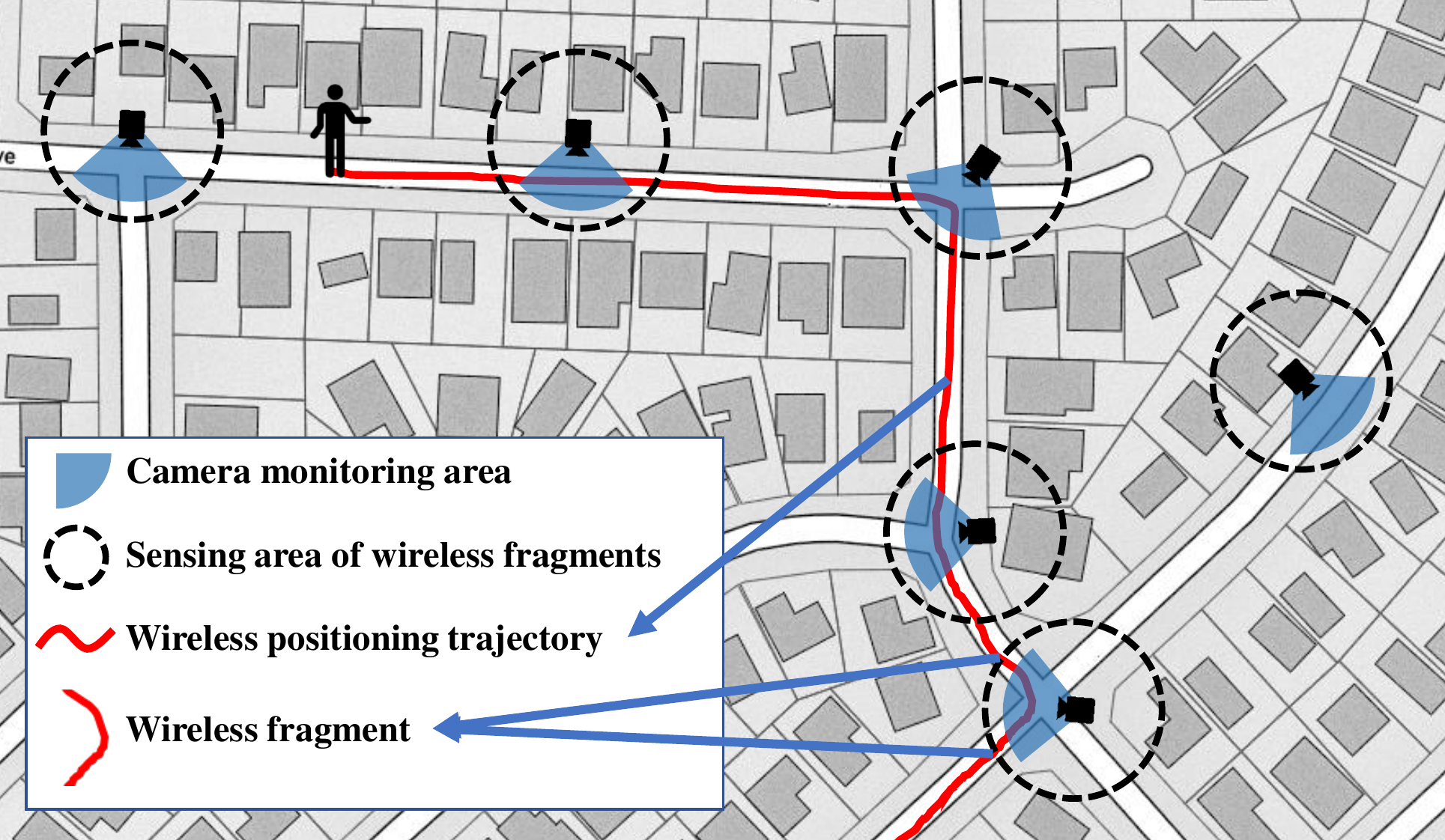}
   \end{center}
   \caption{The problem we study in this work. The wireless positioning trajectories of pedestrians carrying mobile phones can be obtained by existing cellular networks and WiFi positioning. Video data are captured when pedestrians walk to the monitoring areas of cameras. 
   We consider the area within a preset sensing radius centered on the camera as the sensing area of the wireless trajectory fragments.
   When a pedestrian carrying a mobile phone enters the sensing area of the wireless fragments, the fragment of the wireless positioning trajectory within the sensing area is the sensed wireless fragment.
   The wireless fragment may belong to the same pedestrian as one of the videos captured by this camera during its time range.}
   \label{fig:task}
   \end{figure}

   The existing unsupervised person re-identification methods rely on visual cues for model training.
   Although remarkable progress has been made, the defects of the visual data limit their further improvements. 
   In other words, the occlusion and blur lead to the loss of distinguishing parts of pedestrians. Besides, the changes in viewpoints and clothing cause significant appearance changes in pedestrians. 
   These visual noises can easily mislead existing methods that rely solely on visual clues. 
   When the visual data is unreliable, the performance of these methods cannot be guaranteed.
   These defects of visual data force us to seek new supplementary information to increase the robustness of the system.
   
   For the person re-identification task, there have been some attempts to use the multimodal data. Fan \emph{et al}. \cite{fan2020learning} propose to use radar signals for supervised person re-identification, but additional radar equipment is required.
   Liu \emph{et al}. \cite{liu2020vision} introduce the wireless position trajectories for cross-domain person re-identification. Wireless positioning trajectories can be obtained through off-the-shelf equipment, such as cellular networks and WiFi positioning systems. 
   The immutability of signal ID allows it to assist vision-based person re-identification task.
   However, they need to label the GPS coordinate of each location of the monitoring areas to map the pixel coordinate to the world coordinate for calculating the distance between videos and wireless signals \cite{liu2020vision}. 
   Such a paradigm suffers the scalability issue since in real scenarios, labeling and maintaining the location information of each position in all monitoring areas extremely consumes human labor.
   
\begin{figure}
    \begin{center}
    \includegraphics[width=0.95\linewidth]{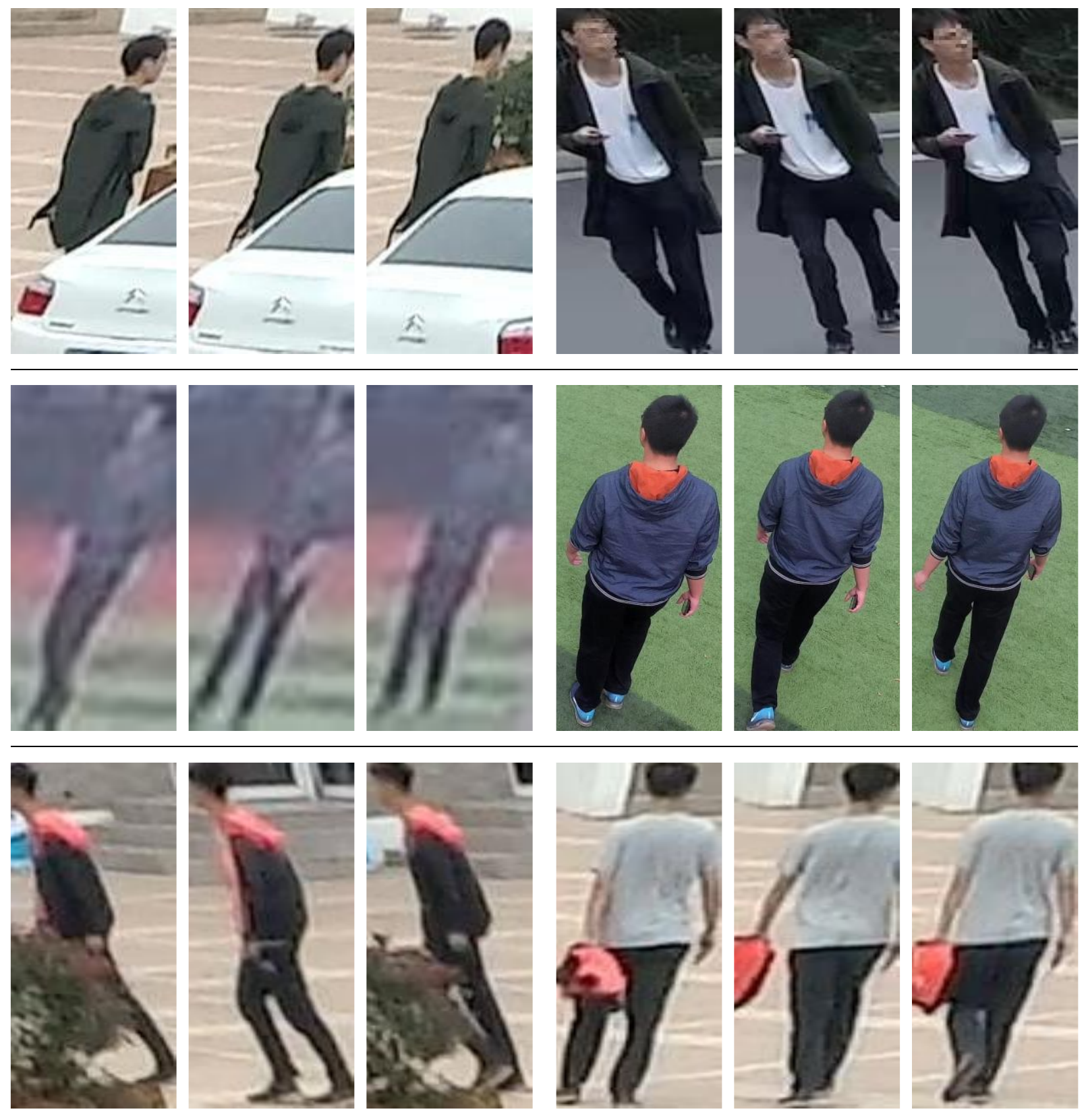}
    \end{center}
    \caption{The occlusion, blur and clothing changes examples on existing person re-identification datasets, which introduce many challenges for existing methods that rely on visual data. Each row in the figure contains two video sequences belonging to the same person, each with three frames.}
    \label{fig:person}
    \end{figure}

   Based on the above discussion, as shown in Fig.~\ref{fig:task}, we propose a new setting to assist unsupervised person re-identification with wireless trajectories to mitigate the effects of visual noise (Fig.~\ref{fig:person}). 
   For pedestrians carrying mobile phones, their wireless positioning trajectories can be obtained through existing cellular networks and WiFi positioning.
   We define the circular area whose distance from the surveillance camera is less than a preset distance as the wireless fragment sensing area of this camera.
   Once a pedestrian carrying a mobile phone enters this area, we assume that a person enters the monitored area and a wireless fragment is sensed. The wireless fragment is the fragment of the wireless positioning trajectory within the circular sensing area.
   The video of the owner of the wireless trajectory should belong to the set of videos captured by the corresponding camera during the time range of this wireless fragment.
   This weak relationship allows us to use wireless data to assist person re-identification.
   Different from \cite{fan2020learning}, our setting does not require new equipment. 
   Compared with~\cite{liu2020vision} which assumes the GPS coordinate labeling on the entire monitoring areas, our weak scene labeling setting only needs to access the locations of cameras. Therefore, our setting is more feasible and scalable for large-scale surveillance scenarios than previous settings \cite{fan2020learning, liu2020vision}. 

   Our weak scene labeling makes the monitoring system easier to maintain and deploy, but it also brings great challenges. In each surveillance scene, since many pedestrians are carrying mobile phones, multiple videos and multiple wireless fragments are captured simultaneously. 
   For a wireless fragment, the video of its owner is mixed in the videos captured by the corresponding camera during its time range, but we don't know which video corresponds to, since the association between videos and wireless fragments is unknown. 
   This weak physical connection brings great challenges for us to use it to assist person re-identification. 
   
   In this work, we propose a new task, which is to use wireless positioning trajectories to assist unsupervised person re-identification under weak scene labeling.
   To handle the challenges in this new task, we devise a novel unsupervised multimodal training framework (UMTF), which contains a multimodal data association strategy (MMDA) and a multimodal graph neural network (MMGN). MMDA constructs detailed multimodal data associations through an adaptive clustering method. 
   MMGN passes multimodal messages in the video graph by adaptively learning the adjacency matrix from multimodal data.
   By using these modules to mine the potential clues in unlabeled multimodal data, UMTF improves the model quality progressively.
   The experimental results conducted on two challenging datasets, \emph{i.e.} WP-ReID and Campus4K, demonstrate the effectiveness of our method.

   \section{Related Works}\label{sec:relatedWrok}
   
   In this section, we first briefly review the progress of supervised and unsupervised person re-identification. Then we introduce some research efforts with  multimodal data to assist vision tasks, especially person re-identification.
   
   \subsection{Person Re-identification}
   
   \textbf{Supervised Person Re-Identification.}
   To solve the difficulties caused by occlusion, blur, dramatic variations of pedestrian's postures and viewpoints, a lot of work are devoted to designing better models to extract distinguishable and robust visual feature representation.
    According to the type of data processed, these methods can be divided into image-based methods \cite{qian2017multi, suh2018part, liu2017hydraplus, ye2021deep, hou2021feature, liu2021end} and video-based methods \cite{wang2016person, chung2017two, liu2019spatial, liu2017quality, li2020relation, yan2020learning}.

   For image-based person re-identification, in \cite{zhao2017deeply}, a part-aligned representation model is proposed to handle the body misalignment problem. 
   The human body is decomposed into regions that are weighted with the part maps generated by a part net. 
   In \cite{qian2017multi}, a multi-scale deep learning model is designed to learn multi-scale discriminative feature representations and a saliency-based learning fusion layer is used to automatically determine the most suitable scales for matching. 
   In \cite{liu2017hydraplus}, HydraPlus-Net is proposed to capture multiple attentions from low-level to semantic-level and extract the multi-scale attention-strengthened features. 
   Liu \emph{et al}. \cite{liu2021end} use both person ID and camera ID to train a model to decouple features into the foreground and background features.
   
   Video-based person re-identification \cite{mclaughlin2016recurrent, chung2017two, liu2019spatial, zhang2020ordered, gou2018systematic} mainly focuses on integrating temporal information. McLaughlin \emph{et al}. \cite{mclaughlin2016recurrent} introduce a recurrent neural network to fuse the temporal features extracted from each frame. 
   Chung \emph{et al}. \cite{chung2017two} use a siamese network to extract features from the original video frames and optical flow images, respectively. 
   The motion information contained in the optical flow is integrated into the final features.
   Liu \emph{et al}. \cite{liu2019spatial} design a recurrent refining unit to integrate appearance representations and motion information to suppress the noise in the videos and refine the features.
   In \cite{liu2017quality, li2020relation}, different kinds of attention networks are designed to fuse the features of video frames.
   
   \textbf{Unsupervised Person Re-Identification.} To reduce the cost of manual labeling and increase the scalability in real scenarios, unsupervised person re-identification aims to train a discriminative model on unlabeled person re-identification datasets. 
   Due to the lack of training labels, how to mine the potential relationships among data is the key to this task.
   As a representative solution, Li \emph{et al}. \cite{li2019unsupervised} propose an unsupervised tracklet association learning (UTAL) model to automatically eliminate the pedestrian ID labeling. 
   Through the per-camera tracklet discrimination learning and cross-camera tracklet association learning, UTAL model achieves good pedestrian distinguishing capability. 
   Chen \emph{et al}. \cite{chen2018deep} jointly optimize two margin-based association losses to constrain the association of each frame and automatically discover the more reliable cross-camera tracklets. 
   In \cite{ye2017dynamic, wu2019unsupervised}, the dynamic graph matching (DGM) method is used to iteratively update the
   data graph and the label estimation process. With the intermediate estimated labels. the model can learn a better feature representation. Xie \emph{et al}. \cite{xie2020progressive} introduce the spatio-temporal constraint to promote the label generation. 
   Lin \emph{et al}. \cite{lin2020unsupervised} introduce several auxiliary information as additional priors for constraints, including a camera-based term that is useful for distance amendment.
   
   \subsection{Multimodal Data and Vision Tasks}
   
   \begin{figure*}[t]
    \begin{center}
    \includegraphics[width=0.95\linewidth]{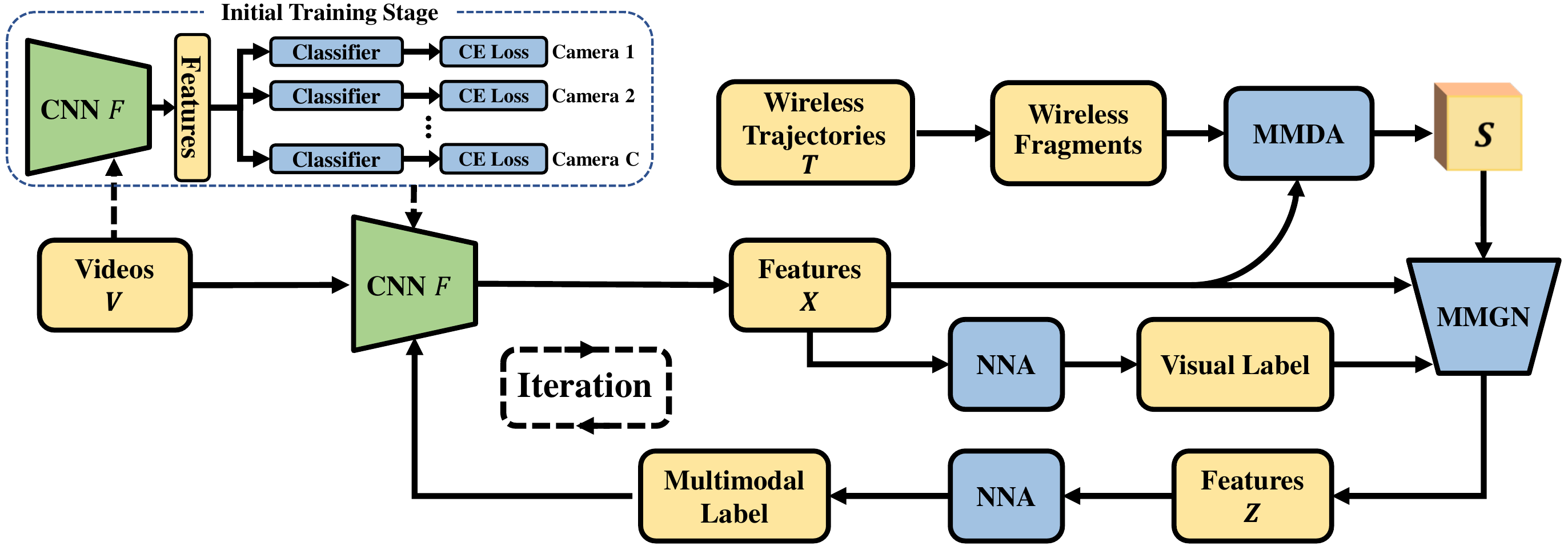}
    \end{center}
    \caption{The overall workflow of our unsupervised multimodal training framework (UMTF). 
    NNA is the nearest neighbor association. MMDA is the multimodal data association strategy. MMGN means the multimodal graph neural network.
    The CNN model obtained from the initial training stage is sent to the second training stage to be alternately trained with MMGN. 
    The CNN model and MMGN generate pseudo visual labels and pseudo multimodal labels to guide and promote each other in the multiple iterations and updates. 
    }
    \label{fig:overview}
    \end{figure*}
   
   Since visual data is easily affected by blur, occlusion, and background clutter, \emph{etc.}, seeking multimodal data to make up for the defects of visual data has become an emerging research direction. To this end, some settings \cite{alahi2015rgb, papaioannou2015accurate, korany2019xmodal, lu2019autonomous, lu2020nowhere} have been explored with promising results.
   
   \textbf{WiFi Signals and Vision Tasks.}
   In \cite{alahi2015rgb, papaioannou2015accurate}, the wireless signal strength is used to assist visual tracking tasks. 
   But the change of signal strength is very susceptible to the influence of the environment, which makes their method unreliable.
   Korany \emph{et al}. \cite{korany2019xmodal} extract gait information from the CSI magnitude measurement of WiFi for person identification, which requires a high quality of visual data to facilitate the human 3D mesh extraction.
   
   \textbf{WiFi Sniffing and Face Recognition.}
   In \cite{lu2020nowhere, lu2019autonomous}, WiFi sniffing is adopted to sense the wireless signal of a person's mobile device for face recognition.
   However, for privacy protection, the current mobile phone system returns random MAC addresses for WiFi sniffing, which does not support the assumption of this approach. Meanwhile, these methods need to divide the data into different events such as meals, meetings, and fitness according to time and location. Since the person re-identification task is uninterrupted and the content is consistent, this strategy does not apply to person re-identification.
   
   \textbf{Multimodal Data and Person Re-Identification.} 
   There are also some efforts \cite{fan2020learning, liu2020vision} to explore the use of multimodal data on person re-identification.
   Fan \emph{et al}. \cite{fan2020learning} utilize radar signals for person re-identification without considering visual data. Liu \emph{et al}. \cite{liu2020vision} introduce the wireless positioning trajectories of mobile devices to assist cross-domain person re-identification. 
   To measure the distance between the videos and the wireless trajectories, this method needs to manually label the GPS coordinate of each location under each camera scene to map the pixel coordinate to the world coordinate. 
   However, it is laborious to label and maintain the coordinates of all positions, which limits their application in large-scale surveillance scenarios.

   \textbf{Privacy Protection of Wireless Data.}
   The wireless positioning trajectories of the mobile phones carried by pedestrians are important privacy data for pedestrians. At present, there are mainly two ways to obtain such data without the permission of pedestrians: WiFi positioning and cellular network positioning. If a pedestrian is connected to WiFi devices provided by someone with unknown intentions, the positioning information of the pedestrian will be leaked. An effective way to avoid this is that pedestrians should only select trusted WiFi devices to avoid the leakage of location data. For the cellular network positioning, the location data is provided by the mobile communication company, which makes it easy to be protected and regulated by relevant organizations to avoid abuse and leakage of data.
   
   \section{Our Method}\label{sec:ourMethod}
   Fig.~\ref{fig:overview} shows the overall workflow of our proposed unsupervised multimodal training framework (UMTF), which mainly includes two training stages. The first stage performs intra-camera model learning to obtain a model $F(\cdot)$ with basic pedestrian discrimination capability.
   In the second training stage, $F(\cdot)$ and the multimodal graph neural network (MMGN) generate pseudo visual labels and pseudo multimodal labels to guide each other's training.
   In the following subsections, after introducing our problem formulation, we will elaborate UMTF from three aspects: the generation of pseudo visual labels, the generation of pseudo multimodal labels, and the mutual promotion of the dual models.

   \subsection{Problem Formulation}

   In Fig.~\ref{fig:overview}, the input of UMTF contains two kinds of data, \emph{i.e.} the video sequences $\{\mathbf{V}_i\}_{i=1}^N$ captured by $C$ cameras and the wireless positioning trajectories $\{\mathbf{T}_m\}_{m=1}^M$. 
   $N$ is the number of video sequences, $C$ is the total number of cameras, and $M$ is the number of wireless trajectories.
   Under the setting of our weak scene labeling, we know the location of each camera.
   As shown in Fig.~\ref{fig:task}, for each camera, a circular area with a preset radius around it is considered as its wireless sensing area.
   The part of the wireless trajectory $\mathbf{T}_m$ located inside the circular wireless fragment sensing area is treated as a wireless fragment $\mathbf{T}_m^r$.
   Every time the owner of the wireless trajectory passes through a monitoring area, a wireless fragment is sensed.
   $\{\mathbf{T}_m^r\}_{r=1}^{R_m}$ denotes $R_m$ wireless fragments belonging to wireless trajectory $\mathbf{T}_m$ and they share the same signal ID, such as MAC address.

   The features of video sequences $\{\mathbf{V}_i\}_{i=1}^N$ are fine-grained descriptions of the appearance of pedestrians but are sensitive to visual noise. Meanwhile, it is unknown which video sequences belong to the same person. The wireless trajectories $\{\mathbf{T}_m\}_{m=1}^M$ are very reliable, but only roughly describe the movement of pedestrians.
   The associations between video sequences and wireless fragments are also unknown. We don't know which video sequence and which wireless fragment belong to the same person. In our proposed new task, \emph{i.e.} unsupervised person re-identification with wireless positioning under weak scene labeling, we only label the locations of cameras. The goal of this task is to train a high-performance person re-identification model $F(\cdot)$ using unlabeled visual data and unlabeled wireless positioning trajectories when knowing the locations of cameras.

   \begin{table}[t]
      \caption{The definition of different mathematical notations in our method.}
      \label{tab:notation}
        \begin{center}
        \begin{tabularx}{0.48\textwidth}{Sl|X}
            \hline
            Variable & Definition \ \\
            \hline
             $\mathbf{V}_i$ & The $i^\mathrm{th}$ video sequence. \\
             $\mathbf{T}_m$ & The $m^\mathrm{th}$ wireless positioning trajectory. \\
             $\mathbf{T}_m^r$ &  The $r^\mathrm{th}$ wireless fragment of $\mathbf{T}_m$. \\
             $R_m$                  & The total number of wireless fragments belonging to $\mathbf{T}_m$.  \\
             $\mathcal{V}_m^r$ & The video set related to $\mathbf{T}_m^r$ \\
             $\mathbb{C}_{m, k}$ & The $k^\mathrm{th}$ cluster. \\
             $K_m$                & The number of clusters of videos related to $\mathbf{T}_m$. \\
             $\mathbf{P}_{m, k}$  & The probability of cluster $\mathbb{C}_{m, k}$ belonging to $\mathbf{T}_m$. \\
             $\mathbf{S}_{i, j, m}$  & The wireless similarity measurement between $\mathbf{V}_i$ and $\mathbf{V}_j$ under $m^\mathrm{th}$ wireless trajectory. \\
             $\mathbf{A}^{\mathrm{avg}}$ & The average result of the wireless channel of $\mathbf{S}$.  \\
             $\mathbf{S}^{\mathrm{h}}$ & The histogram of the wireless channel of $\mathbf{S}$.  \\
             $\mathbf{A}$         & The final adjacency matrix of video graph. \\
            \hline
          \end{tabularx}
        \end{center}
        \end{table}

   
   
   \subsection{Generation of Pseudo Visual Labels}
   \label{method:gpvl}
   
   \textbf{Initial Training Stage.} To obtain the pseudo visual labels, we need an initial model with basic pedestrian discrimination capability.
   As shown in the initial training stage of Fig.~\ref{fig:overview}, we use a multi-branch network architecture \cite{li2019unsupervised, xie2020progressive} to train the basic model. 
   The network has a shared feature extraction model $F(\cdot)$ and $C$ independent classifiers. 
   Given a video sequence $\mathbf{V}_i$, the model $F(\cdot)$ extracts the feature of each frame and averages them to get the video-level feature $\mathbf{x}_i\in {\mathbb{R}}^{2048}$.
   The features of the video sequences from each of $C$ cameras are fed to the corresponding classifier. For each classifier, the number of classes is equal to the number of video sequences captured by the corresponding camera, \emph{i.e.} every video sequence is treated as an independent class. We use the cross-entropy loss to train each classifier. The total loss is the sum of the $C$ cross-entropy losses.
   
   \textbf{Nearest Neighbor Association.} 
   After the initial training stage, we obtain a model $F(\cdot)$ with basic pedestrian discrimination capability.
   Given $N$ video sequences, we extract their features with $F(\cdot)$.
   Then, we apply the nearest neighbor association (NNA) \cite{li2019unsupervised, xie2020progressive} for cross-camera video sequences association and obtain the pseudo visual labels.
   Specifically, for each video sequence, NNA first finds the cross-camera nearest neighbor on each camera pair based on the cosine similarity of the features.
   Then, for two video sequences, if they are the cross-camera nearest neighbor of each other, they are regarded as belonging to the same person and assigned the same pseudo visual label.
   Since being the cross-camera nearest neighbors to each other is a strong constraint, not all video sequences are assigned pseudo visual labels.
   During the training process, we filter out these unlabeled video sequences, but this does not affect the design of the algorithm.
   Therefore, in the following narration, we assume that all video sequences are assigned pseudo visual labels for brevity.
   
   \subsection{Generation of Pseudo Multimodal Labels}
   The generation of pseudo multimodal labels relies on two main modules, \emph{i.e.}, multimodal data association strategy (MMDA) and multimodal graph neural network (MMGN). MMDA associates visual data and wireless data with the adaptive clustering method. Different from GCN \cite{kipf2016semi} and GAT \cite{velivckovic2017graph}, MMGN adaptively learns the adjacency matrix from the histogram statistics of wireless similarity.
   
   \textbf{Multimodal Data Association.}
   In the setting of \cite{liu2020vision}, since the latitude and longitude of each location in scenes captured by cameras is labeled, the visual data and wireless data are directly correlated by measuring the distance between the visual trajectories and the wireless trajectories. However, in our task, we adopt the weak scene labeling. We only need to know the locations of cameras, not the latitude and longitude of each location of the scenes captured by the cameras. This reduces the cost of scene annotation, but it also brings great challenges to the association and fusion of multimodal data. To this end, we design a multimodal data association strategy (MMDA), which divides the visual data and wireless data in time and space based on the locations of cameras, and then mine the correlation between them.

   \begin{figure}[t]
      \begin{center}
      \includegraphics[width=0.99\linewidth]{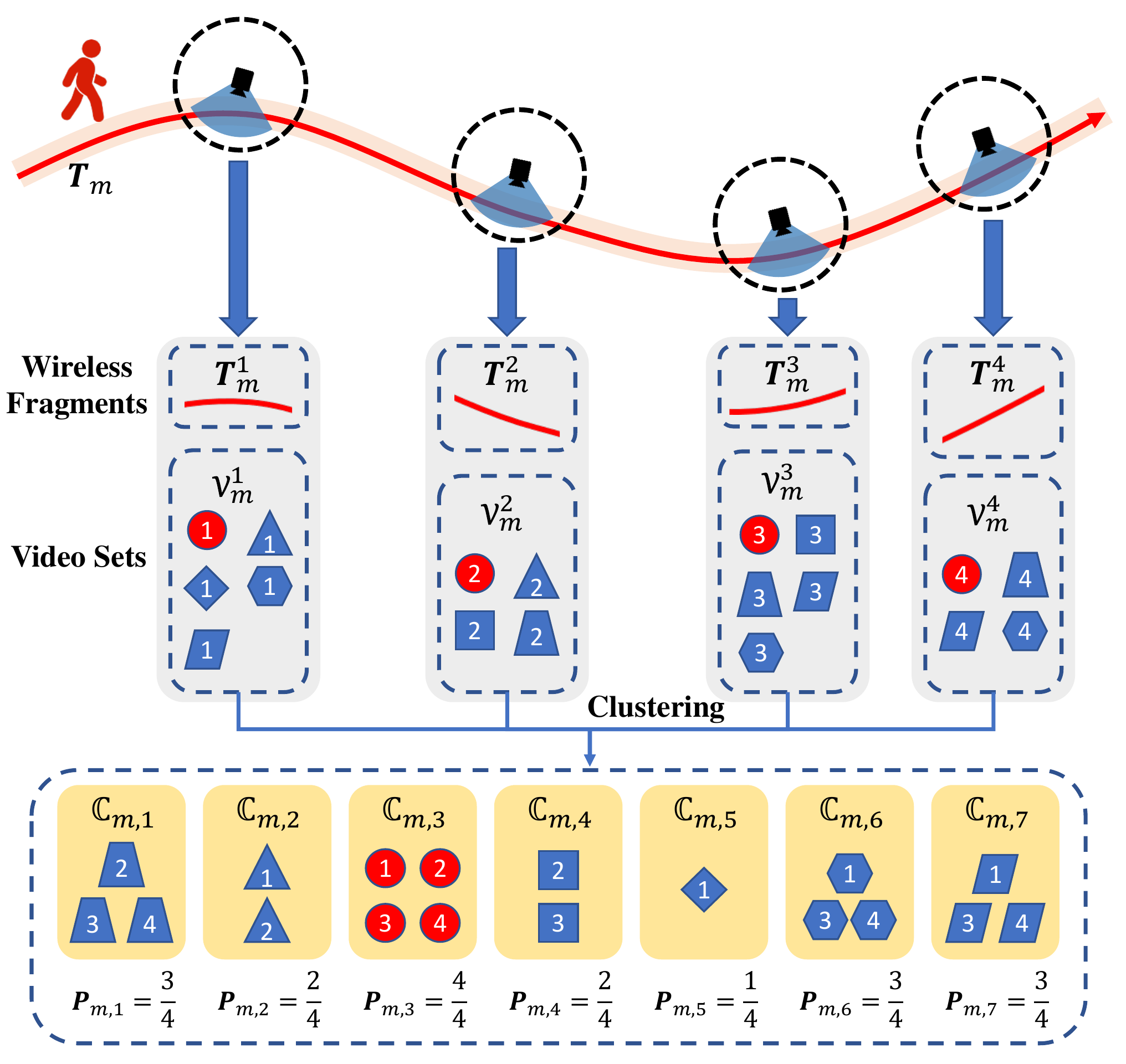}
      \end{center}
      \caption{The process of multimodal data association (MMDA). The circular area (the area inside the dotted circle) with a preset radius around each camera is regarded as the wireless fragment sensing area. 
      The part of a pedestrian's wireless positioning trajectory $\mathbf{T}_m$ located in a wireless sensing area is regarded as a wireless fragment $\mathbf{T}_m^r$.
      In this example, 4 wireless fragments $\{\mathbf{T}_m^r\}_{r=1}^{4}$ are sensed when the pedestrian pass by 4 surveillance cameras.
      \emph{We use different markers to represent video sequences. Video sequences with the same marker share the same person identity.
      The number in a marker is the serial number of the corresponding wireless fragment.}
      Taking $\mathbf{T}_m^1$ as an example, during the time range of $\mathbf{T}_m^1$, 5 video sequences of pedestrians are captured by the corresponding camera, which are denoted as a video set $\mathcal{V}_m^1$. 
      The video sequence (red circle) of the target pedestrian is mixed in the video set $\mathcal{V}_m^1$.
      MMDA is dedicated to correctly associating the wireless trajectory of the target person with its video sequences (red circles).
      }
      \label{fig:MMDA}
      \end{figure}

   As shown in Fig.~\ref{fig:MMDA}, a circular area (the area inside the dotted circle) with a preset radius around a surveillance camera is defined as the sensing area for wireless fragments.
   For a target pedestrian carrying a mobile phone, we can obtain the wireless positioning trajectory $\mathbf{T}_m$ by cellular networks or WiFi positioning. 
   The part of the wireless trajectory $\mathbf{T}_m$ located in a wireless sensing area is the sensed wireless fragment $\mathbf{T}_m^r$.
   We use different markers to represent video sequences. Video sequences with the same marker share the same person identity.
   The number in a marker is the serial number of the corresponding wireless fragment.
   During the time range of the wireless fragment $\mathbf{T}_m^1$, if there is only the target pedestrian in the surveillance area, the video sequences captured by the camera at this time belongs to the same person as the wireless fragment. However, many pedestrians often appear in the monitoring scene at the same time. In this case, the video sequence (red circle) of the target person is mixed with a bunch of video sequences $\mathcal{V}_m^1$ captured by the corresponding camera during the time range of this wireless fragment $\mathbf{T}_m^1$. We use $\mathcal{V}_m^r$ to represent these video sequences related to wireless fragment $\mathbf{T}_m^r$. 
   
   In Fig.~\ref{fig:MMDA}, given multiple wireless fragments $\{\mathbf{T}_m^r\}_{r=1}^{4}$, we obtain corresponding related video sets $\{\mathcal{V}_m^r\}_{r=1}^{4}$.
   The video sequences (red circles) belonging to the target pedestrian are mixed with these video sets. The goal of MMDA is to accurately estimate the probability that each video sequence in $\{\mathcal{V}_m^r\}_{r=1}^{4}$ belongs to the same person as this wireless trajectory $\mathbf{T}_m$. 

   Since the person re-identification model $F(\cdot)$ has a certain capability to distinguish pedestrians, we use $F(\cdot)$ to extract the features of the video sequences in $\{\mathcal{V}_m^r\}_{r=1}^{4}$. As shown in Fig.~\ref{fig:MMDA}, we apply the $k$-means clustering algorithm on the features and get $K_m$ clusters $\{\mathbb{C}_{m, k}\}_{k=1}^{K_m}$. 
   The value  of $K_m$, \emph{i.e.}, the number of the clusters is estimated adaptively, which will be described in a later paragraph.
   Video sequences belonging to the same pedestrian are similar and their features are likely to be clustered together. Therefore, we assume that the video sequences in the same cluster belong to the same person. This assumption introduces some noise because the clustering results are not always accurate. However, our experimental results show that with the increase of training epochs, the clustering results will get better and better and achieve satisfactory results.  This assumption can still assist multimodal data association well.  

   Since a wireless fragment $\mathbf{T}_m^r$ is sensed when wireless trajectory $\mathbf{T}_m$ passes through a monitoring area. As illustrated in Fig.~\ref{fig:MMDA}, $\mathbf{T}_m$ has $R_m = 4$ wireless fragments $\{\mathbf{T}_m^r\}_{r=1}^{R_m=4}$ because it passes through 4 monitoring areas. Therefore, the owner of this wireless trajectory should have passed through these four monitoring areas as well. 
   This indicates that we can use the consistency of the motion path of the pedestrian represented by the video sequences contained in the cluster $\mathbb{C}_{m, k}$ and the path of the wireless trajectory $\mathbf{T}_m$ to measure the possibility that they belong to the same person.  

   For each video sequence, we know which camera it is captured by. So for the pedestrian represented by video cluster $\mathbb{C}_{m, k}$, we can easily obtain the number of monitoring scenes it passes through. We define function $Q(\mathbb{C}_{m, k})$ to return this number. Taking the clustering results in Fig.~\ref{fig:MMDA} as an example, the video sequences contained in the cluster $\mathbb{C}_{m, 7}$ are from the video sets $\mathcal{V}_m^1$, $\mathcal{V}_m^3$ and $\mathcal{V}_m^4$. Therefore, $Q(\mathbb{C}_{m, 7}) = 3$. This indicates that the pedestrians represented by $\mathbb{C}_{m, 7}$ passed through three of the $R_m = 4$ monitoring areas that the wireless trajectory $\mathbf{T}_m$ passed through. Their paths are not perfectly matched. To measure the probability that the video cluster $\mathbb{C}_{m, k}$ and the wireless trajectory $\mathbf{T}_m$ belong to the same person, we define the consistency between the pedestrian's motion path represented by the video cluster $\mathbb{C}_{m, k}$ and that of the wireless trajectory $\mathbf{T}_m$ as  
   \begin{equation}
      \label{equ:pro:signal}
      \mathbf{P}_{m, k} = \frac{Q(\mathbb{C}_{m, k})}{  R_m  } \,.
   \end{equation}
   The closer $\mathbf{P}_{m, k}$ is to 1, the more consistent the path of the pedestrian represented by cluster $\mathbb{C}_{m, k}$ is with the path of the wireless trajectory $\mathbf{T}_m$, and the more likely they are to belong to the same person.

   When the wireless trajectory passes through the same monitoring scene many times, the number of scenes does not increase, which causes the number of scenes to not reflect the pedestrian's path well. To this end, when the pedestrian's wireless trajectory passes through the same monitoring area again at different times, MMDA will perceive a new wireless fragment and specially mark this monitoring scene, instead of treating it as the same monitoring scene.

   For the cluster $\mathbb{C}_{m, 3}$ in Fig.~\ref{fig:MMDA}, $Q(\mathbb{C}_{m, 3}) = 4$. The pedestrian represented by it passes through all $R_m = 4$ monitoring areas where the wireless trajectory $\mathbf{T}_m$ passes, so $\mathbf{P}_{m, 3} = \frac{Q(\mathbb{C}_{m, 3})}{  R_m  } = \frac{4}{4} = 1$, which indicates that the video sequences in this cluster are likely to belong to the same person as the wireless trajectory $\mathbf{T}_m$.
   So far, through the processing of MMDA shown in Fig.~\ref{fig:MMDA}, we obtain the association between the visual data and the wireless trajectory $\mathbf{T}_m$. By repeating this process for all wireless trajectories, we can obtain the association between the visual data and all wireless data.

   \textbf{Adaptive estimation of the number of clustering centers.}
   In the above process of obtaining associations between video sequences and the wireless trajectory $\mathbf{T}_m$, we need to apply $k$-means algorithm on video sets $\{\mathcal{V}_m^r\}_{r=1}^{R_m}$. Instead of using a manually preset $K_m$ for the number of clustering centers, we design a method to automatically estimate the appropriate number for $K_m$. 
   
   Generally, every time a pedestrian passes through a monitoring scene of a camera, a video sequence and a wireless fragment are sensed. Therefore, the average number of wireless fragments of a trajectory can roughly reflect the average number of video sequences belonging to a person, which can be calculated as $\frac{\sum\nolimits_{m=1}^{M}R_m}{M}$. $R_m$ is the number of wireless fragments belonging to $m^\mathrm{th}$ wireless trajectory $\mathbf{T}_m$.  $\sum\nolimits_{m=1}^{M}R_m$ is the total number of wireless fragments of all wireless trajectories. $M$ is the total number of wireless trajectories. $R_m$ and $M$ are known values that do not require human labeling.
   $\frac{\sum\nolimits_{m=1}^{M}R_m}{M}$ reflects the average number of wireless fragments a wireless trajectory (pedestrian) has, and is also an appropriate estimate of the average number of video sequences a pedestrian has. 

   For wireless trajectory $\mathbf{T}_m$, the video sequences in video sets $\{\mathcal{V}_m^r\}_{r=1}^{R_m}$ are related to it. The total number of these video sequences related to $\mathbf{T}_m$ is $ \sum\nolimits_{r=1}^{R_m} \lvert \mathcal{V}_m^r \rvert $.
   Since we have estimated that the average number of video sequences a pedestrian has is $\frac{\sum\nolimits_{m=1}^{M}R_m}{M}$.
   Then, we can estimate that these video sequences related to wireless trajectory $\mathbf{T}_m$ come from $\frac{ \sum\nolimits_{r=1}^{R_m} \lvert \mathcal{V}_m^r \rvert \times M}{\sum\nolimits_{m=1}^{M}R_m}$ pedestrians. 
   
   Considering the purpose of clustering is to gather videos belonging to the same person into one cluster, $\frac{ \sum\nolimits_{r=1}^{R_m} \lvert \mathcal{V}_m^r \rvert \times M}{\sum\nolimits_{m=1}^{M}R_m}$ is a good estimate for the number of clustering centers $K_m$.
   However, in a real situation, a pedestrian passing through a monitoring scene may have intermittent tracking videos due to occlusion, which makes the pedestrian has one wireless fragment but be captured in multiple video sequences in this scene. This makes the number of wireless fragments a person owns not a completely accurate reflection of the number of video sequences a person owns. Besides, the walking routes of pedestrians are different, which makes the estimated person number deviating from the true person number. 
   Considering the above cases, we amend the estimation formula of $K_m$ as
   \begin{equation}
      \label{equ:cluster:k}
      K_m = \lambda\frac{ \sum\nolimits_{r=1}^{R_m} \lvert \mathcal{V}_m^r \rvert \times M}{\sum\nolimits_{m=1}^{M}R_m}\,,
   \end{equation}
   where coefficient $\lambda$ controls the proportional relationship between $K_m$ and the estimated number of pedestrians. 
   By roughly adjusting $\lambda$ to an appropriate value, $K_m$ can approach the true person number very well.
   Instead of manually selecting the same number $K$ for all wireless trajectories without clues, our method can adaptively calculate the appropriate number of clusters for videos released to different trajectories.

   It is worth emphasizing that the above strategy for estimating the number of clustering centers does not break the unsupervised task setting. Each wireless trajectory has a unique identification, \emph{e.g.}, MAC address, so we can directly obtain the number of wireless trajectories without manual labeling. wireless fragments are obtained directly by judging the distance between the wireless trajectories and cameras, which are also free of manual labeling. The videos $\{\mathcal{V}_m^r\}_{r=1}^{R_m}$ related to a pedestrian's wireless trajectory $\mathbf{T}_m$ are a small fraction of the videos captured by the entire surveillance network. We cannot know exactly how many pedestrians these videos are from without manually labeling. Therefore, we roughly estimate it by estimating the number of video sequences a pedestrian has. This estimated number, although not completely accurate, is an appropriate reference for the number of clustering centers. 

   \textbf{Wireless Similarity.} Although we obtain the association between visual data and wireless data by MMDA shown in Fig.~\ref{fig:MMDA}, the association results of the multimodal data need to be further processed to be used in the training of person re-identification models.
   For two video sequences in the cluster $\mathbb{C}_{m, k}$, when $\mathbf{P}_{m, k}$ is large, both of them are likely to belong to one pedestrian. This means that the two video sequences should be very similar, which suggests that we can take this value as the similarity of the two videos based on wireless data.
   With such consideration, we define a similarity measurement strategy between $\mathbf{V}_i$ and $\mathbf{V}_j$ based on $m^\mathrm{th}$ wireless trajectory as
   \begin{equation}
      \label{equ:aff:ws}
      \mathbf{S}_{i, j, m} \!= \!\left\{
      \begin{array}{l}
      \!\!\mathbf{P}_{m, k}  \quad \!  \mathrm{if} \,  i \neq j  \,   \mathrm{and} \,  \exists \mathbb{C}_{m,k}, \{\mathbf{V}_i, \mathbf{V}_j\} \subseteq \mathbb{C}_{m,k} \,,\\
      \!\!1 \quad \quad \   \,   \mathrm{if} \, i = j \,, \\
      \!\!0 \quad \quad \    \,   \mathrm{otherwise} \,,
      \end{array} \right.
      \end{equation}
      where $\mathbf{S} \in {\mathbb{R}}^{N \times N \times M}$ and $\mathbf{S}_{m} \in {\mathbb{R}}^{N \times N}$ embodies the similarities between the $N$ videos based on the $m^\mathrm{th}$ wireless trajectory $\mathbf{T}_m$.
   During the data association process of MMDA, only video sequences with temporal and spatial overlap with the wireless fragments are related to a corresponding wireless trajectory, which results in $\mathbf{S}_{m}$ being a sparse matrix.
   
    \begin{figure}[t]
      \begin{center}
      \includegraphics[width=0.98\linewidth]{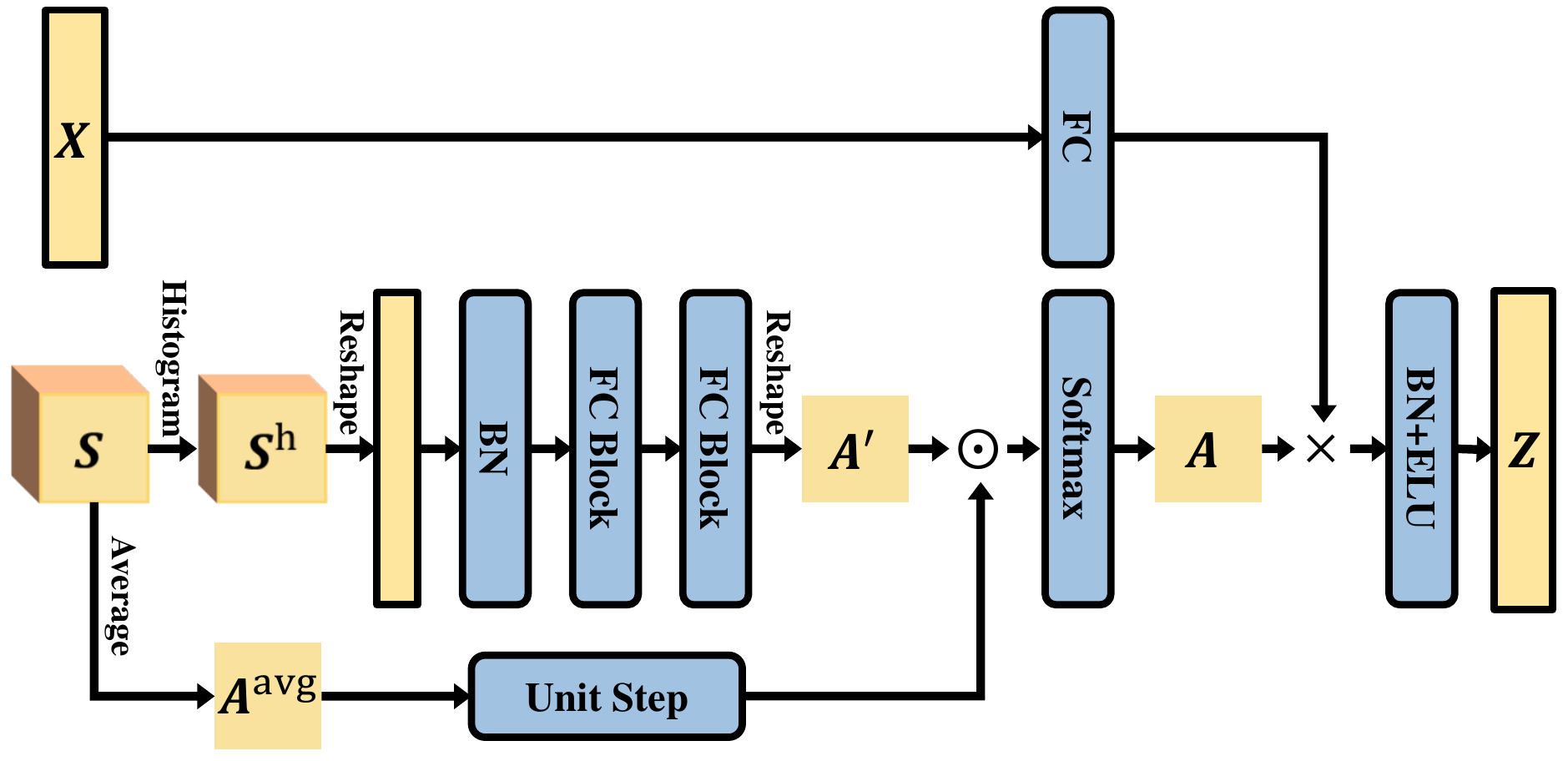}
      \end{center}
      \caption{The architecture of the multimodal graph convolutional module (MGM). $\mathbf{S} \in {\mathbb{R}}^{N \times N \times M}$ are the wireless similarity between $N$ videos. $\mathbf{X} \in {\mathbb{R}}^{N \times 2048}$ are the features of $N$ video sequences. The output features $\mathbf{Z} \in {\mathbb{R}}^{N \times 512}$ are the video features updated with wireless information.
      $\odot$ denotes the element-wise multiplication. $\times$ is the matrix multiplication.}
      \label{fig:MGM}
      \end{figure}

      \begin{figure}[t]
        \begin{center}
        \includegraphics[width=0.75\linewidth]{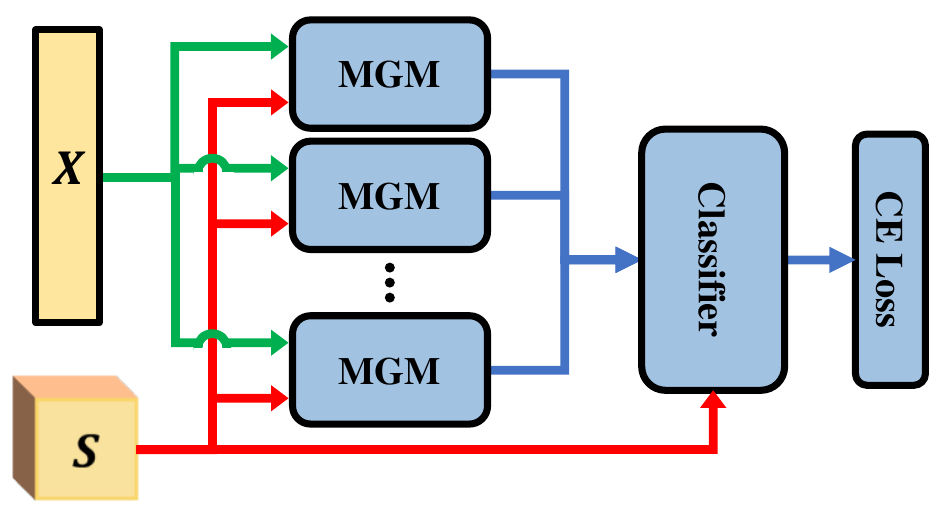}
        \end{center}
        \caption{The overall architecture of the multimodal graph neural network (MMGN). MGM denotes the multimodal graph convolutional module. $\mathbf{S} \in {\mathbb{R}}^{N \times N \times M}$ are the wireless similarity between $N$ videos. $\mathbf{X} \in {\mathbb{R}}^{N \times 2048}$ are the features of $N$ video sequences.}
        \label{fig:mmgn}
        \end{figure}

   \textbf{Multimodal Graph Neural Network.}
   Through the process described above, we get a new wireless similarity measurement $\mathbf{S}$ between video sequences. To make better use of it, we build a video graph containing $N$ nodes. Each node denotes a video sequence and is described by the feature $\mathbf{x}_i$ obtained from model $F(\cdot)$. After defining the nodes, one way to construct the adjacency matrix is to directly take the mean value of $\mathbf{S}$ along the third dimension as the adjacency matrix $\mathbf{A}^{\mathrm{avg}}$, where $\mathbf{A}^{\mathrm{avg}}_{i,j} = \sum_{m}\mathbf{S}_{i, j, m}/M$. With features $\mathbf{X} \in {\mathbb{R}}^{N \times 2048}$ and the adjacency matrix $\mathbf{A}^{\mathrm{avg}} \in {\mathbb{R}}^{N \times N}$, we can use the graph convolutional network (GCN) \cite{kipf2016semi} to pass message. However, $\mathbf{A}^{\mathrm{avg}}$ is obtained by directly averaging the wireless channel of the sparse matrix $\mathbf{S}$, which will cause oversmoothing if $M$ is very large. Besides, the averaging of the wireless channel loses the distribution information of wireless similarity under different wireless trajectories.

   Based on the above considerations, we propose a multimodal graph convolutional module (MGM) to make better use of the information provided by wireless similarity $\mathbf{S}$. As shown in Fig.~\ref{fig:MGM}, given $\mathbf{S}$, we first count its histogram $\mathbf{S}^{\mathrm{h}} \in {\mathbb{R}}^{N \times N \times 32}$, which is formulated as 
   \begin{equation}
      \label{equ:histogram}
      \mathbf{S}^{\mathrm{h}}_{i,j} = \frac{h(\mathbf{S}_{i,j})}{M} \,,
   \end{equation}
   where $h(\cdot)$ divides the $[0, 1]$ into 32 bins with equal size and returns the histogram of the input. $\mathbf{S}^{\mathrm{h}}_{i,j}$ uses 32 bins to reflect the distribution in wireless similarity $\mathbf{S}_{i,j} \in {\mathbb{R}}^{M}$ between $\mathbf{V}_i$ and $\mathbf{V}_j$ on multiple wireless trajectories. 
   Given the histogram distribution $\mathbf{S}^{\mathrm{h}}$ of wireless similarity $\mathbf{S}$, we further use several neural network layers to adaptively learn the adjacency matrix $\mathbf{A} \in {\mathbb{R}}^{N \times N} $ of the graph.

   \begin{figure*}
    \begin{center}
    \includegraphics[width=0.8\linewidth]{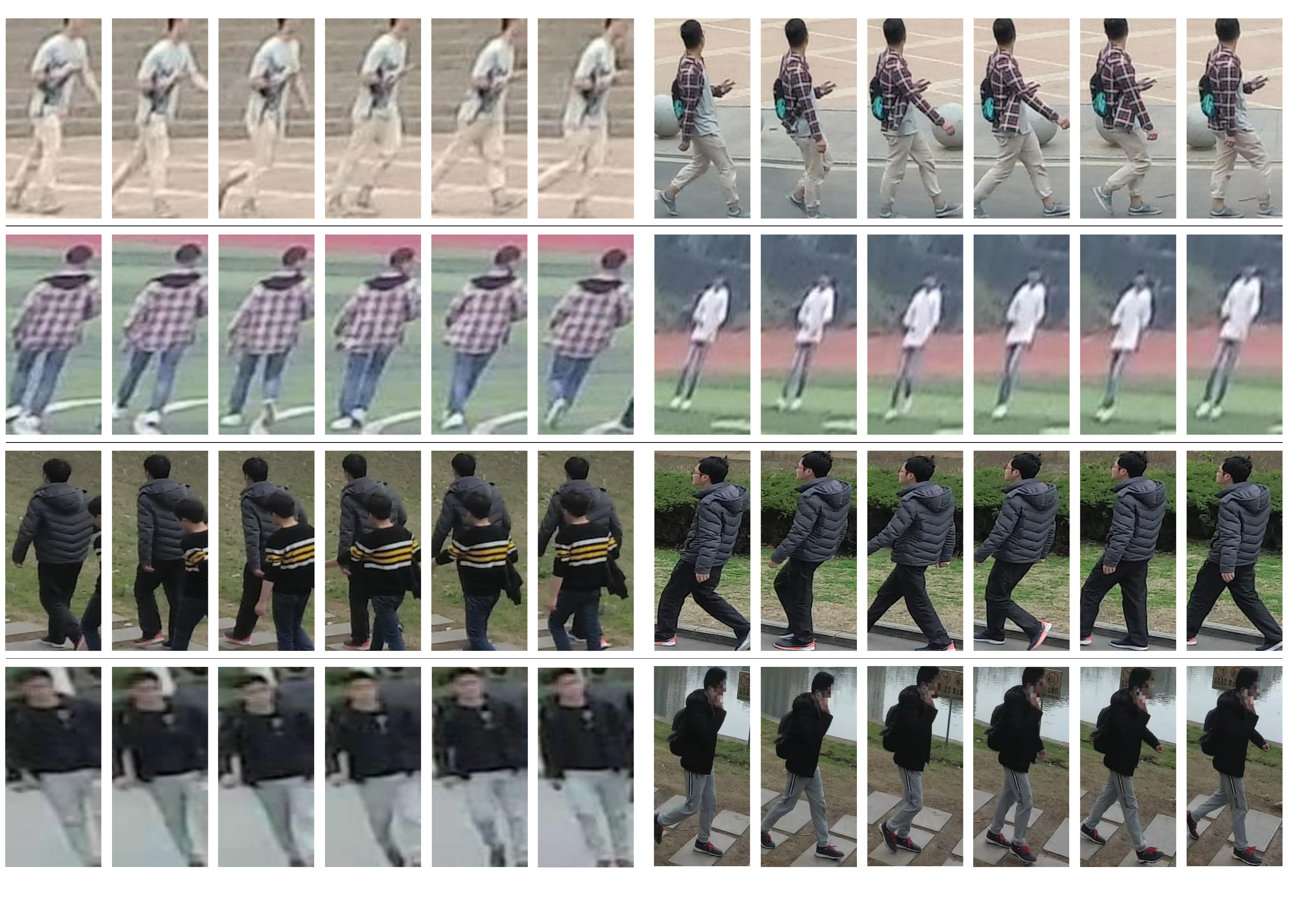}
    \end{center}
    \caption{Examples of the matching results of our method. Each row in the figure contains two videos belonging to the same person, each with six frames. The first two rows of videos are from WP-ReID dataset, and the videos in the next two rows are from Campus4K. These videos are difficult to match correctly by existing vision-based methods because of clothes changing, blur and occlusion issues. Thanks to the reliability of the wireless signals, our method avoids the interference of visual factors and matches them correctly.}
    \label{fig:example}
    \end{figure*}
   
   As shown in Fig.~\ref{fig:MGM}, $\mathbf{S}^{\mathrm{h}}$ is first reshaped to a 2D matrix with shape ${\mathbb{R}}^{NN \times 32}$ and each 32-dim vector is treated as a feature vector. Then, the result is fed to two FC blocks to get $\mathbf{A}^{\prime} \in {\mathbb{R}}^{N \times N}$. Each block contains a fully connected layer (FC), a batch normalization (BN) layer and a leaky rectified linear unit. 
   These two FC blocks output 16-dim and 1-dim features, respectively.
   Finally, we get the adjacency matrix $\mathbf{A}$ by $\mathbf{A} = \mathrm{softmax}(\mathbf{A}^{\prime} \odot u(\mathbf{A}^{\mathrm{avg}}))$, where $u(\cdot)$ is the unit step function. $u(\cdot) = 1$ when the input value is larger than 0, otherwise $u(\cdot) = 0$. $\odot$ denotes the element-wise multiplication. $\mathbf{A}^{\prime} \odot u(\mathbf{A}^{\mathrm{avg}})$ filters out the learned association between two video sequences in $\mathbf{A}^{\prime}$ that have no wireless-based association. The softmax operation is to normalize each row of the matrix.
   
   Different from GAT \cite{velivckovic2017graph} that learns the adjacency matrix from pairs of video features, our adjacency matrix $\mathbf{A}$ is learned from the wireless similarity that contains multimodal information with our carefully designed network structure. Meanwhile, compared with the adjacency matrix $\mathbf{A}^{\mathrm{avg}}$ obtained by the averaging operation, $\mathbf{A}$ makes better use of the wireless similarity and adapts to this graph.

   After obtaining the learned adjacency matrix $\mathbf{A}$,  we first embed the features $\mathbf{X} \in {\mathbb{R}}^{N \times 2048}$ of $N$ video sequences into 512-dim features by a FC layer. Then, by multiplying $\mathbf{A}$, we get the video features after message passing. Finally, we get the final video features $\mathbf{Z} \in {\mathbb{R}}^{N \times 512}$ after the processing of BN and exponential linear unit (ELU). As shown in Fig.~\ref{fig:MGM}, $\mathbf{Z}$ is the video feature representation after updating the raw video features $\mathbf{X}$ with the wireless similarity $\mathbf{S}$ between videos. Since $\mathbf{Z}$ considers both visual and wireless data, it can also be called the multimodal feature representation of video sequences.
   
   With the designed MGM, as shown in Fig.~\ref{fig:mmgn}, we further propose a multimodal graph neural network (MMGN) to strengthen the use of multimodal data.
   To enhance the representation capability of features, similar to \cite{vaswani2017attention, velivckovic2017graph}, we use the multi-head mechanism to concatenate features generated by 6 MGMs to achieve the final representation. The classifier is also a MGM module, while the final BN layer and ELU layer are removed and the dimension of the output is the number of classes. 
   
   \subsection{The Mutual Promotion of the Dual Models}
   
   In our proposed unsupervised multimodal training framework (UMTF), different modules cooperate to get better results progressively. As shown in Fig.~\ref{fig:overview}, in the initial training stage, we obtain a model $F(\cdot)$ with basic pedestrian discrimination capability. In the second training stage, model $F(\cdot)$ extracts the visual feature representation $\mathbf{X} \in {\mathbb{R}}^{N \times 2048}$. With $\mathbf{X}$, as described in Section \ref{method:gpvl}, we use the nearest neighbor association (NNA) to generate pseudo visual labels. 
   
   On the video graph, MMGN learns the adjacency matrix adaptively from the histogram distribution of wireless similarity $\mathbf{S}$ with the specially designed multimodal graph convolutional modules (MGM). The input of MMGN is the raw videos features $\mathbf{X}$. The output of MMGN is video features mixed with wireless information, \emph{i.e.}, the multimodal features $\mathbf{Z} \in {\mathbb{R}}^{N \times 512}$ of video sequences. MMGN is trained by cross-entropy loss with pseudo visual labels.
   
   With the trained MMGN, we update $\mathbf{X}$ and get updated multimodal features $\mathbf{Z} \in {\mathbb{R}}^{N \times 512}$. Similar to the process described in Section \ref{method:gpvl}, we apply the nearest neighbor association (NNA) on $\mathbf{Z}$ to obtain the pseudo multimodal labels, which can be adopted for the further training of model $F(\cdot)$ with the batch hard triplet loss \cite{hermans2017defense}.
   
   In summary, as illustrated in Fig.~\ref{fig:overview}, the training of $F(\cdot)$ and MMGN are performed alternately. With the pseudo labels provided by the other model, they are trained again to obtain better models and estimate more accurate pseudo labels for the other model. In such multiple iterations, they progressively promote each other and reach a stable state. Finally, we obtain a more discriminative model $F(\cdot)$ for inference.

   \section{Experiment}\label{sec:experiments}
   
   \subsection{Experimental Setup}
   
   \textbf{Datasets.} We evaluate our proposed method on two person re-identification datasets, \emph{i.e.} WP-ReID \cite{liu2020vision} and Campus4K \cite{xie2020progressive}. WP-ReID \cite{liu2020vision} contains both video sequences and authentic wireless positioning trajectories of the mobile phones carried by pedestrians. In WP-ReID, 868 video sequences of 79 pedestrians are captured by six cameras. Among them, 29 pedestrians have corresponding wireless trajectories. The captured scenes of three cameras overlap. Occlusion, blur, and clothing changes are common in this dataset. These factors make WP-ReID very challenging. 
   
   Campus4K \cite{xie2020progressive} is a large person re-identification dataset containing only video data. 3,849 video sequences from 1,567 identities are captured by 6 cameras. 
   Although this dataset has no wireless data, we can generate simulated wireless data that is very close to the real situation.
   We randomly select 80\% pedestrians and assume that they carry mobile phones and can be collected wireless trajectories. 
   For each video sequence $\mathbf{V}_i$ of these people, we can generate a corresponding wireless fragment based on its timestamp. 
   Considering the sensing range of the wireless data is usually larger than the monitoring area of the camera, the time range of the wireless fragment should be larger than the time range of the video sequence. Therefore, we add extra 15 seconds to the beginning and end of the time of the generated wireless fragment to extend the time range by 30 seconds.

   The generation of simulated wireless data relies on the timestamp of each frame of the video sequence in the dataset. 
   In the existing popular video-based large-scale person re-identification dataset, only Campus4K provides timestamps. Therefore, we generate simulated wireless data on Campus4K dataset only. Campus4K is a large-scale dataset with complex background and varied pedestrians, which makes it very challenging. The combination of WP-ReID dataset \cite{liu2020vision} with authentic wireless data and the large-scale Campus4K dataset  \cite{xie2020progressive} with simulated wireless data allows to fully validate the effectiveness of the different methods.

   \begin{table}[t]
    \caption{The ablation study of UMTF. \emph{BL} denotes the baseline method, which directly train the model $F(\cdot)$ with the pseudo visual labels. MMGN$^{\mathrm{Avg}}$ directly takes the mean value of $\mathbf{S}$ as the adjacency matrix $\mathbf{A}^{\mathrm{avg}}$ and 
    uses GCN \cite{kipf2016semi} to pass message. MMGN$^{\mathrm{GAT}}$ uses GAT \cite{velivckovic2017graph} to replace the default MGM in MMGN. When MMGN is adopted, MMDA will also be adopted to provide wireless similarity.}
   \label{tab:gcn:ablation}
   \small
      \begin{center}
      \begin{tabular}{l|cc|cc}
      \hline
      \multirow{2}{*}{Method} & \multicolumn{2}{c|}{WP-ReID} & \multicolumn{2}{c}{Campus4K} \\
      \cline{2-5}
      & mAP & Rank-1 & mAP & Rank-1 \\
      \hline
      Initial Training                            & 37.9 & 76.6 & 74.4 & 72.5  \\
      Baseline                                & 54.7 & 85.6 & 80.8 & 78.8  \\
      BL + MMGN$^{\mathrm{Avg}}$         & 60.0 & 84.6 & 82.5 & 81.0  \\
      BL + MMGN$^{\mathrm{GAT}}$ \cite{velivckovic2017graph} & 61.3 & 87.6 & 84.6 & 82.6  \\
      BL + MMGN (UMTF)                       & \textbf{66.3} & \textbf{91.0} & \textbf{86.6} & \textbf{85.4}  \\
      \hline
      \end{tabular}
      \end{center}
      
      \end{table}
      \begin{figure}[t]
        \begin{center}
        \includegraphics[width=0.7\linewidth]{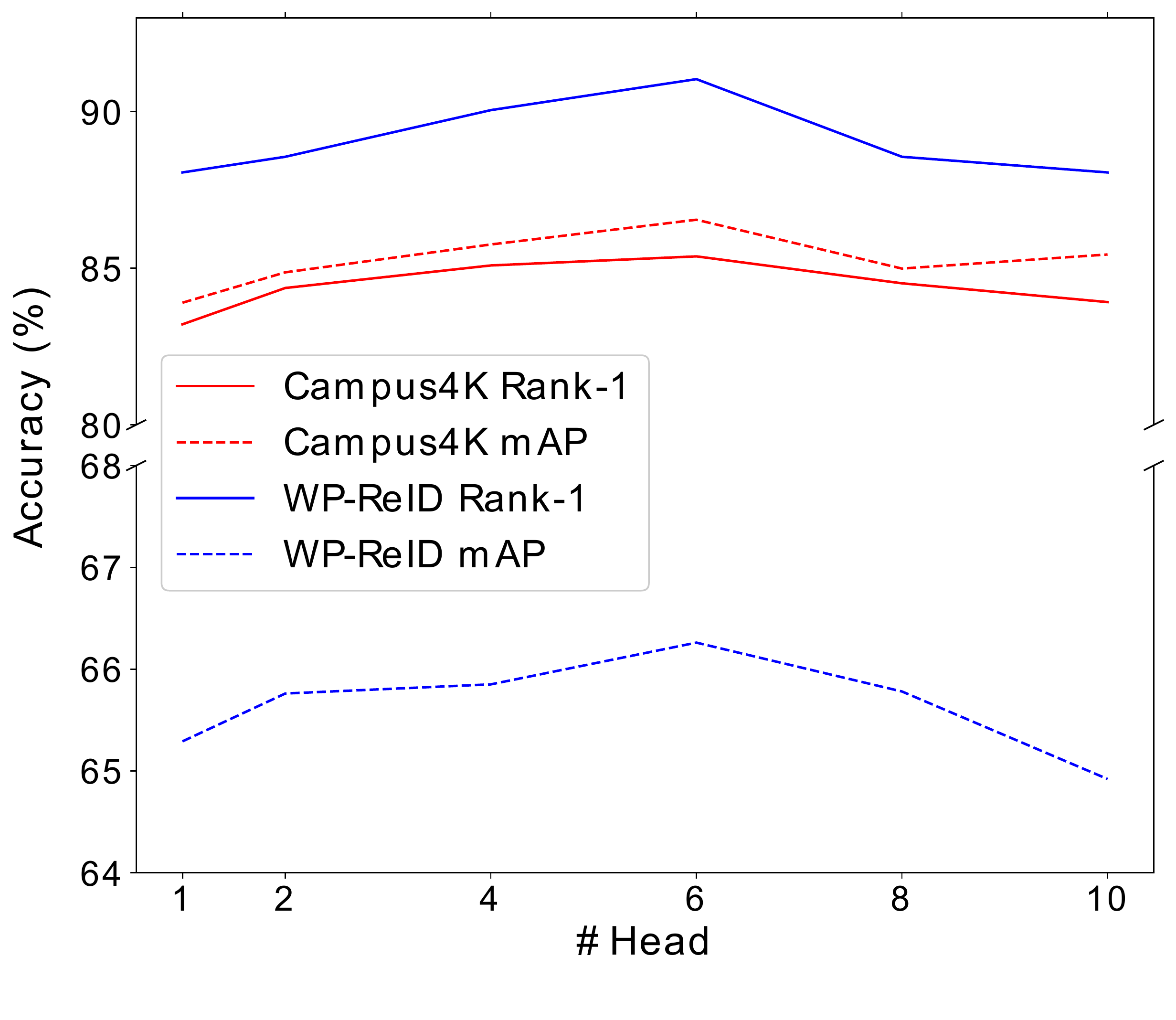}
        \end{center}
        \caption{The effect of the number of heads in the multi-head mechanism of MMGN on the performance of UMTF.}
        \label{fig:head}
        \end{figure}

        \begin{table}[t]
          \caption{The impact of MMGN trained by different combinations of loss functions on the performance of UMTF. CE denotes the cross-entropy loss. Triplet denotes the triplet loss \cite{hermans2017defense}.}
         \label{tab:mmgn:loss}
         \small
            \begin{center}
            \begin{tabular}{l|cc|cc}
            \hline
            \multirow{2}{*}{Method} & \multicolumn{2}{c|}{WP-ReID} & \multicolumn{2}{c}{Campus4K} \\
            \cline{2-5}
            & mAP & Rank-1 & mAP & Rank-1 \\
            \hline
            CE                            & 66.3 & 91.0 & 86.6 & 85.4  \\
            Triplet                       & 58.9 & 85.1 & 83.4 & 82.2  \\
            CE + Triplet                  & 58.8 & 86.0 & 82.9 & 82.2  \\
            \hline
            \end{tabular}
            \end{center}
            \end{table}

   \textbf{Evaluation Protocols.}
   WP-ReID \cite{liu2020vision} only provides test set. Following \cite{liu2020vision}, we directly train the unsupervised model on the test set without any label. For Campus4K \cite{xie2020progressive}, we adopt the standard training and testing split, while the labels of the training set are not used. We take the standard Cumulated Matching Characteristics (CMC) table and mean Average Precision (mAP) as the evaluation metric of the two datasets. For CMC table, the scores at rank 1, 5, 10 are reported. Rank-$r$ visualizes an expectation of finding the correct person in the top $r$ matches. To evaluate the performance of pseudo-label prediction or clustering, we use adjusted mutual information (AMI) as the evaluation metric, which measures the consistency between the obtained results and the ground truth labels.
   
   \textbf{Implementation Details.}
   We take ResNet50 \cite{he2016deep} pretrained on ImageNet as the backbone model. The final FC layer is removed. The input frames are resized to $256 \times 128$ and the number of video frames is 2. For the initial training stage, we update the model for 80 epochs by the Adam algorithm with an initial learning rate of $3\!\times\!10^{-4}$. 
   576 video sequences are randomly selected for each training batch.
   In the second training stage, we re-estimate the pseudo labels every 5 epochs.
   For the training of model $F(\cdot)$ in the second stage, we update the model for 80 epochs by the SGD algorithm with a learning rate of $6\!\times\!10^{-5}$. 
   Random horizontal flipping is adopted for data augmentation.
   We select video sequences of 32 pedestrians each with 16 video sequences as the inputs.
   The margin of the triplet loss is 0.4.
   When training MMGN, the network is updated for 80 epochs by the Adam algorithm with a learning rate of 0.01.
   The weight decay for all model training is set to $5\!\times\!10^{-4}$. 
   For WP-ReID and Campus4K, $\lambda$ is set to 3 and 2 by default, respectively.
   The sensing radius of wireless trajectories in WP-ReID dataset is 50 meters by default. 
   The source code is released at \url{https://github.com/yolomax/UMTF}.

   \begin{figure*}[t]
    \begin{center}
    \includegraphics[width=0.315\linewidth]{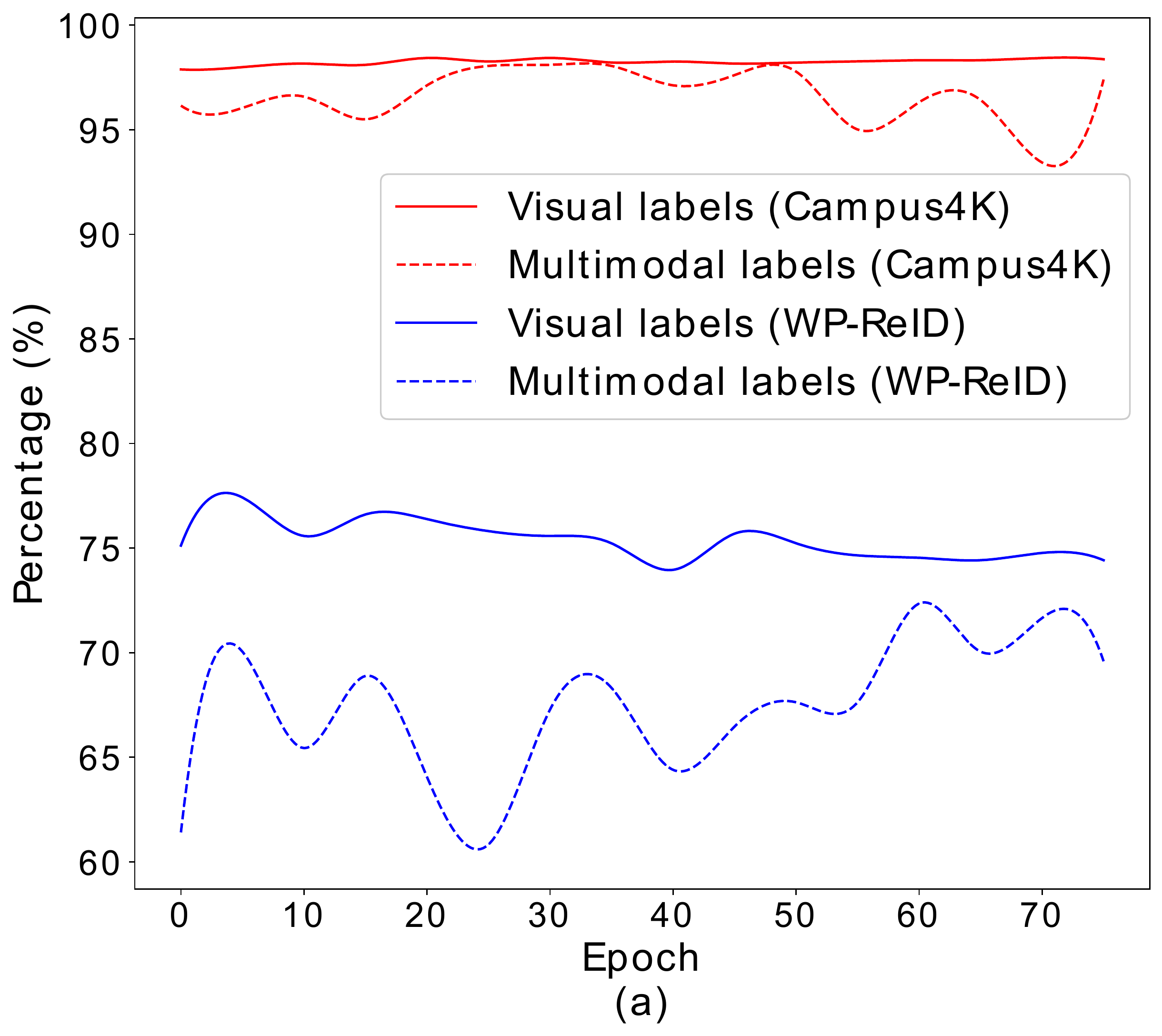}
    \hspace{0.25cm}
    \includegraphics[width=0.315\linewidth]{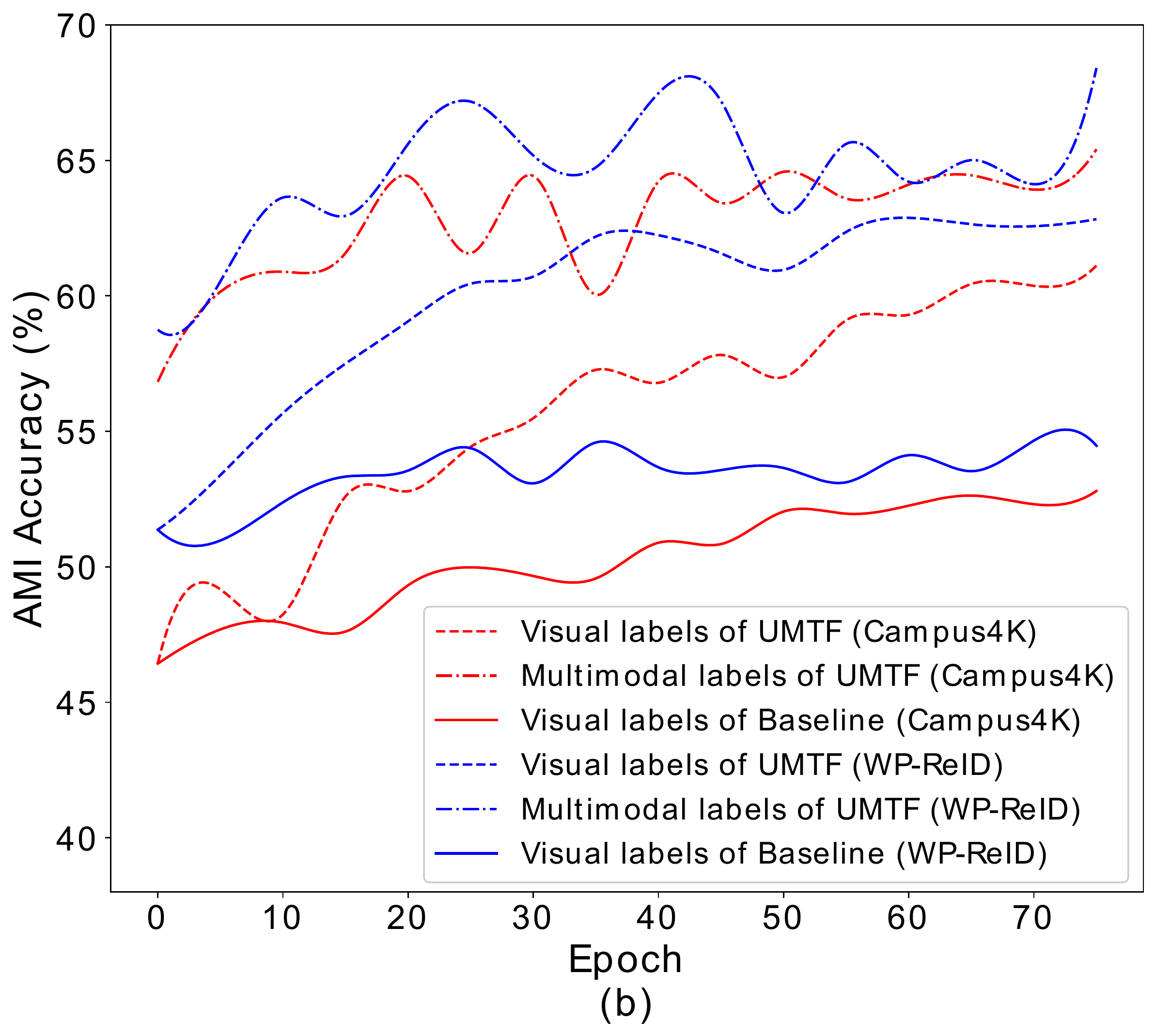}
    \hspace{0.25cm}
    \includegraphics[width=0.315\linewidth]{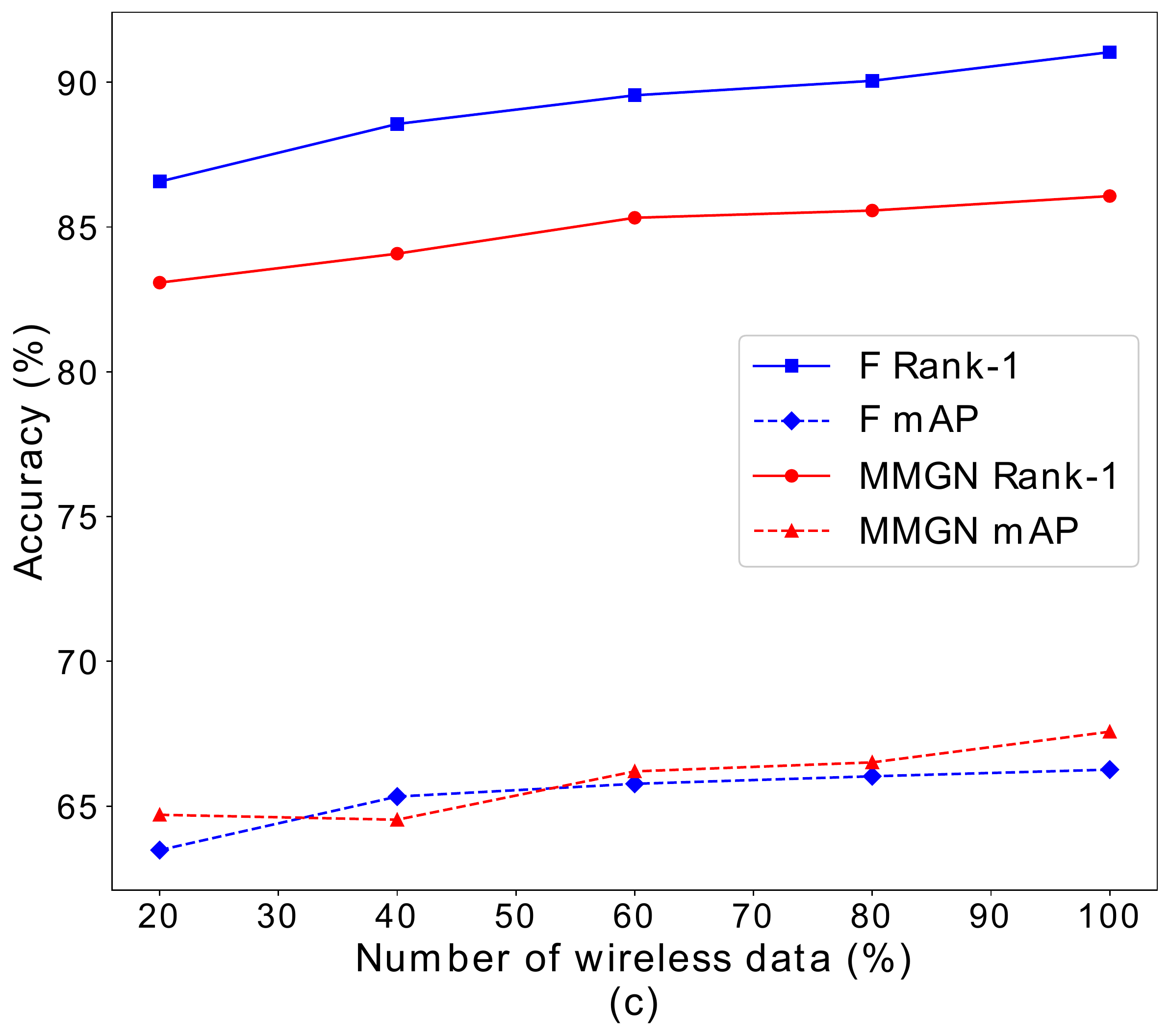}
    \end{center}
    \caption{(a) The number of video sequences assigned pseudo visual labels or pseudo multimodal labels as a percentage of the total number of videos.  
    (b) The prediction accuracies of pseudo labels, which takes adjusted mutual information (AMI) as the evaluation metric.
    (c) The impact of the number of wireless signals on the performance of model $F(\cdot)$ and MMGN on WP-ReID dataset. }
    \label{fig:all:ablation}
    \end{figure*}
   \begin{figure*}[t]
      \begin{center}
      \includegraphics[width=0.315\linewidth]{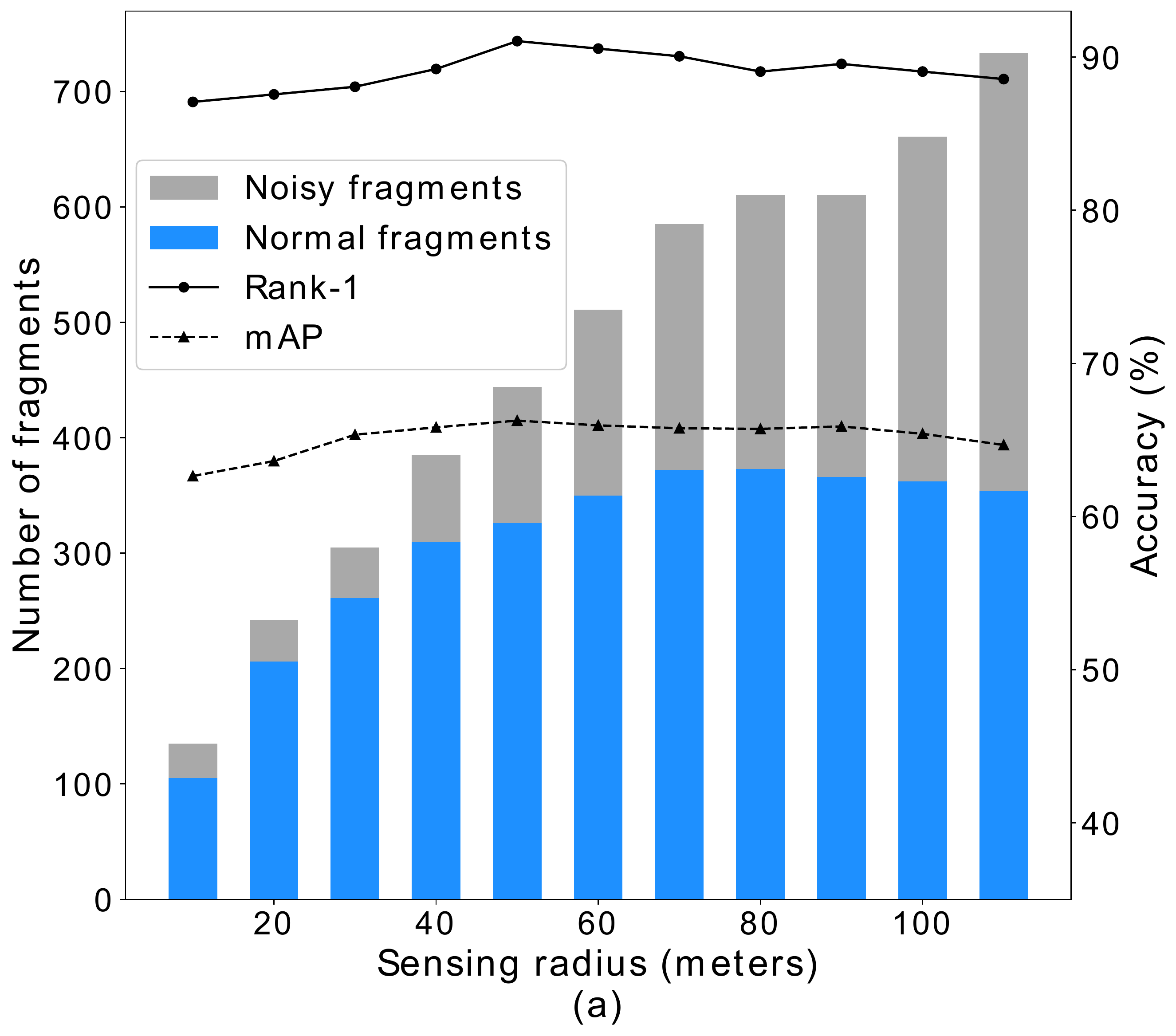}
      \hspace{0.25cm}
      \includegraphics[width=0.315\linewidth]{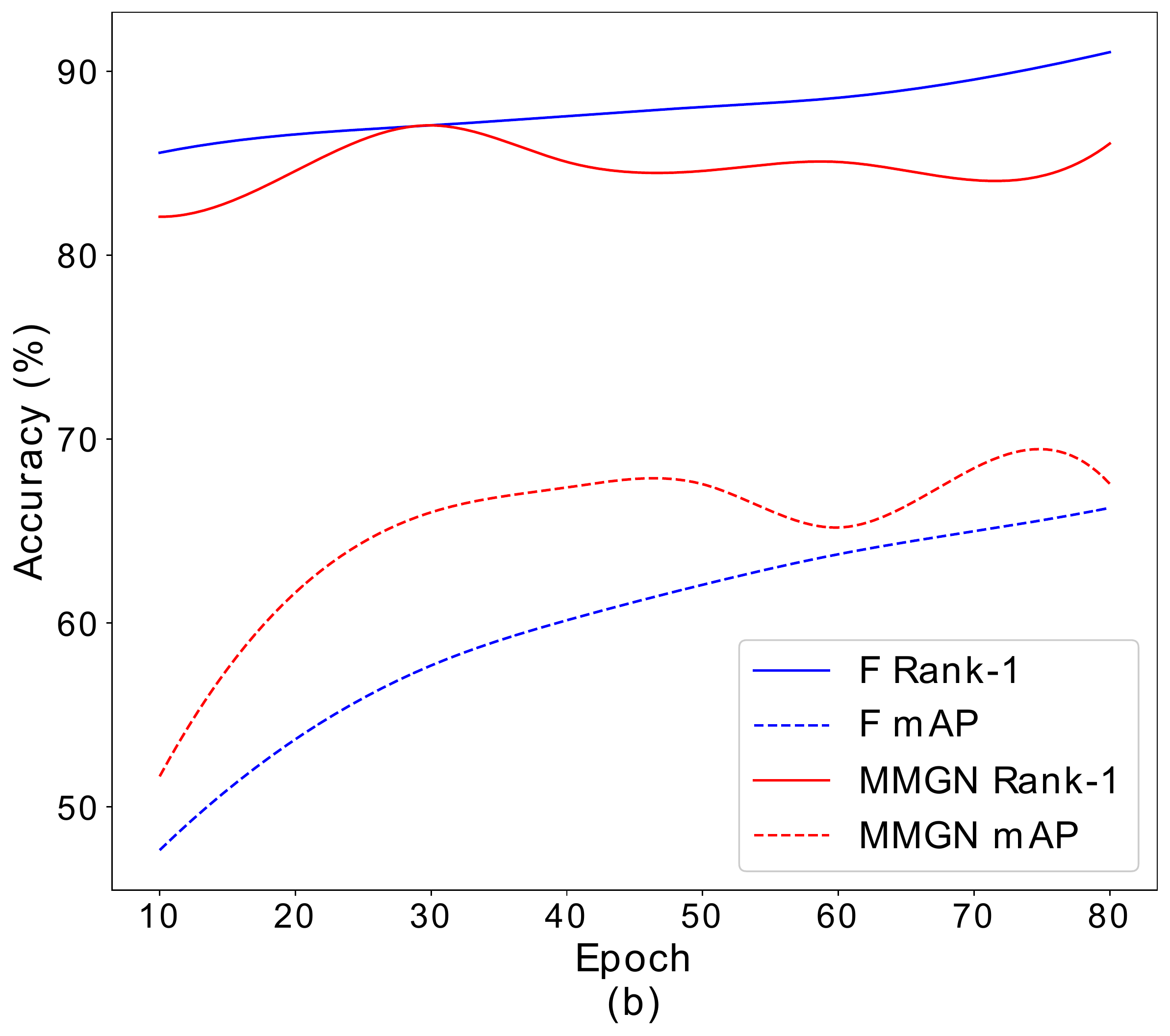}
      \hspace{0.25cm}
      \includegraphics[width=0.315\linewidth]{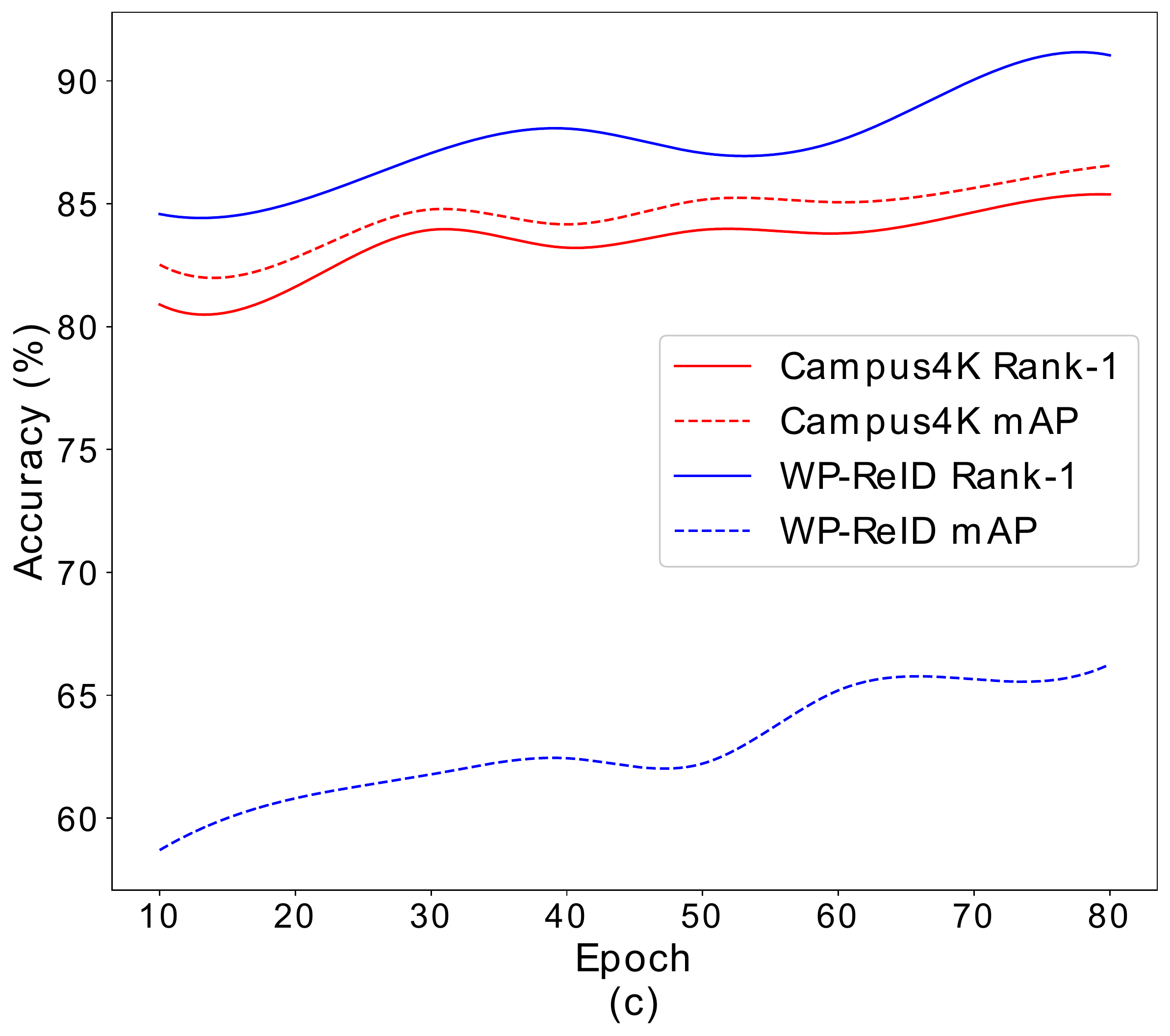}
      \end{center}
      \caption{(a) The influence of the sensing radius of wireless trajectories on the performance of UMTF on WP-ReID dataset. 
      (b) The accuracy changes (\%) of model $F(\cdot)$ and MMGN as the number of training epochs increases on WP-ReID dataset.
      (c) The accuracy changes (\%) of model $F(\cdot)$ as the increase of the epochs MMGN participates in the training in UMTF.
      For example, the value at $\mathrm{Epoch}=50$ is the accuracy of $F(\cdot)$ that trained for 80 epochs, while the MMGN is removed from UMTF after 50th epoch and wireless data is only involved in the first 50 training epoches.}
      \label{fig:utfn:ablation}
      \end{figure*}

   \subsection{Ablation Study}
   \label{exp:ablation}

   As shown in Table~\ref{tab:gcn:ablation}, \emph{Initial Training} directly uses the model obtained after the initial training stage for inference. \emph{Baseline} uses the pseudo visual labels obtained through visual data for CNN model training in the second stage. MMDA and MMGN are not adopted in \emph{Baseline}.
   Neither \emph{Initial Training} nor \emph{Baseline} uses wireless data and their performances are limited. Visual noise makes it difficult for some data to get correct pseudo visual labels to participate in training and causes the final model to fail to correctly match some difficult video pairs (Fig.~\ref{fig:example}).

   In Table~\ref{tab:gcn:ablation}, we evaluate the impact of different ways of obtaining the adjacency matrix on the performance of UMTF. 
   \emph{MMGN$^{\mathrm{Avg}}$} directly takes the mean value of wireless similarity $\mathbf{S}$ as the adjacency matrix $\mathbf{A}^{\mathrm{avg}}$ and uses GCN \cite{kipf2016semi} to pass message. \emph{BL + MMGN$^{\mathrm{GAT}}$} takes $\mathbf{A}^{\mathrm{avg}}$ as the initial adjacency matrix and adaptively learns the new adjacency matrix from video feature pairs by GAT \cite{velivckovic2017graph}. GAT also uses the multi-head strategy. As illustrated in Table~\ref{tab:gcn:ablation}, the performances of \emph{BL + MMGN} using MGM are significantly higher than \emph{BL + MMGN$^{\mathrm{Avg}}$} and \emph{BL + MMGN$^{\mathrm{GAT}}$}. This shows that MGM makes better use of the wireless similarity provided by MMDA. MGM learns the adjacency matrix from the histogram of wireless similarity, which avoids the problem of oversmoothing due to direct averaging of large sparse matrices $\mathbf{S}$, while retaining statistical information on the wireless similarity values under multiple wireless signals.

   In Fig.~\ref{fig:head}, we evaluate the impact of the multi-head mechanism of MMGN on the performance of UMTF. It can be seen that the performance of UMTF is higher when the number of heads increases. The model achieves the best performance on both datasets when the number of heads is 6.

   \begin{figure*}[t]
    \begin{center}
    \includegraphics[width=0.315\linewidth]{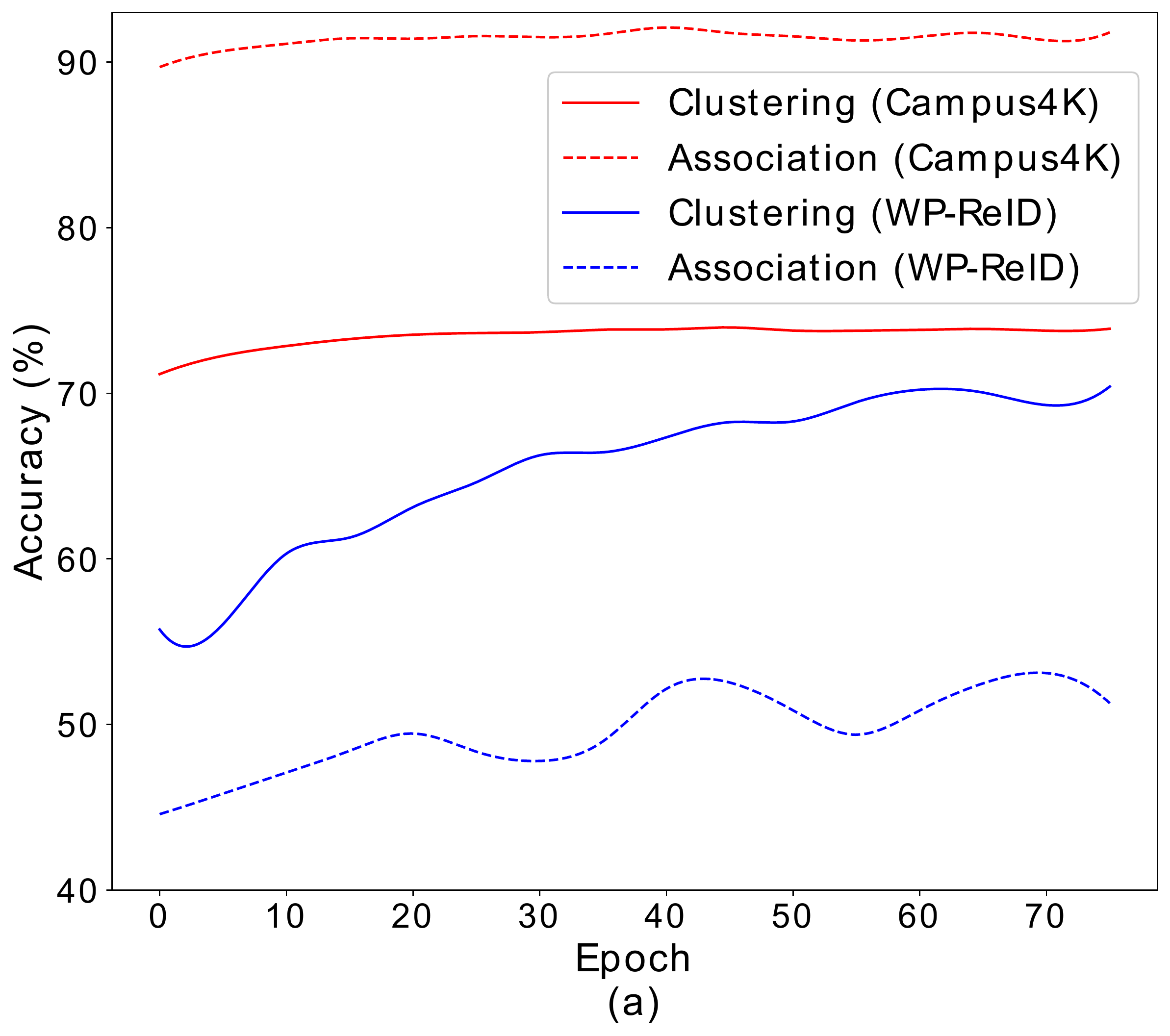}
    \hspace{0.25cm}
    \includegraphics[width=0.315\linewidth]{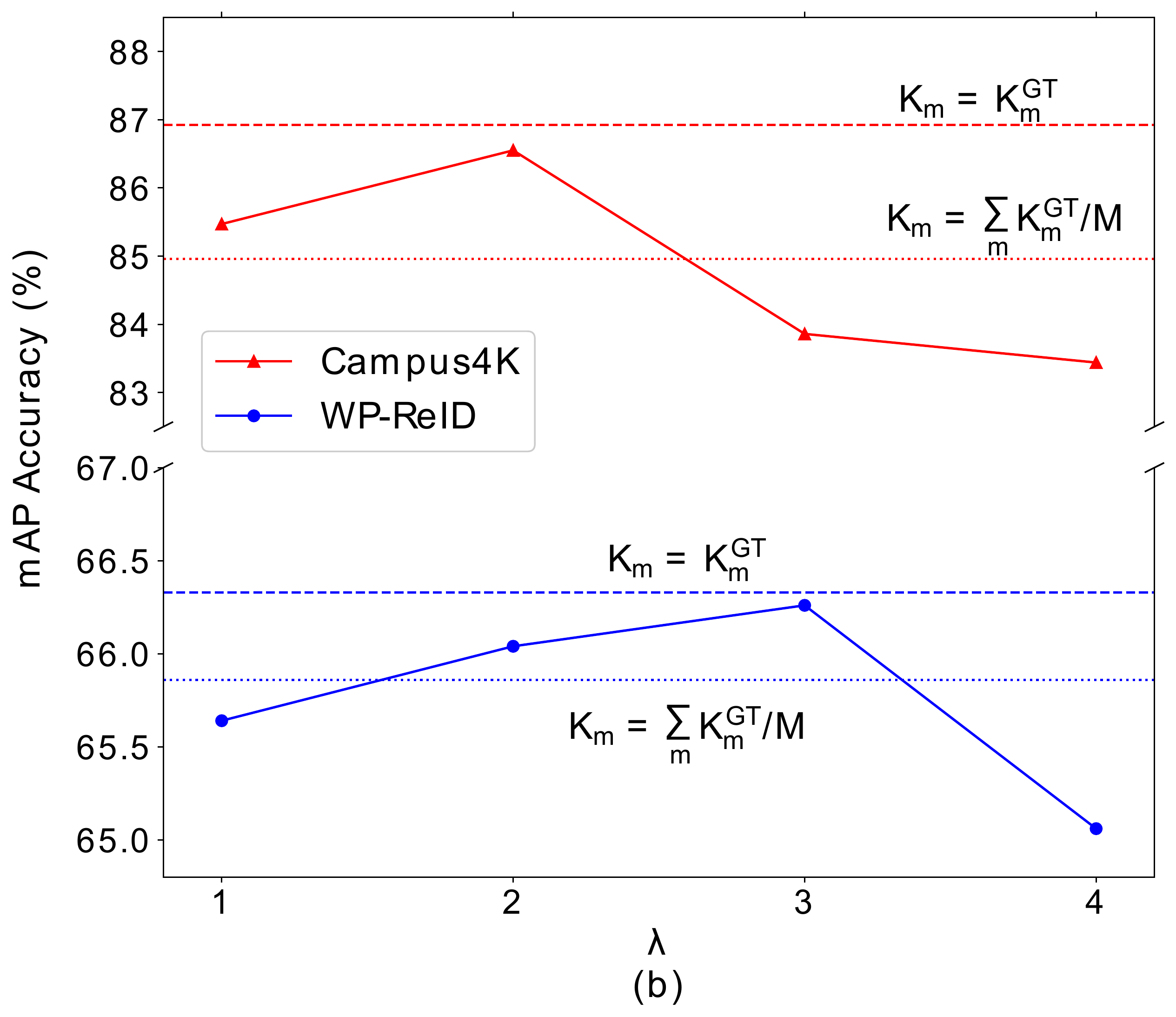}
    \hspace{0.25cm}
    \includegraphics[width=0.315\linewidth]{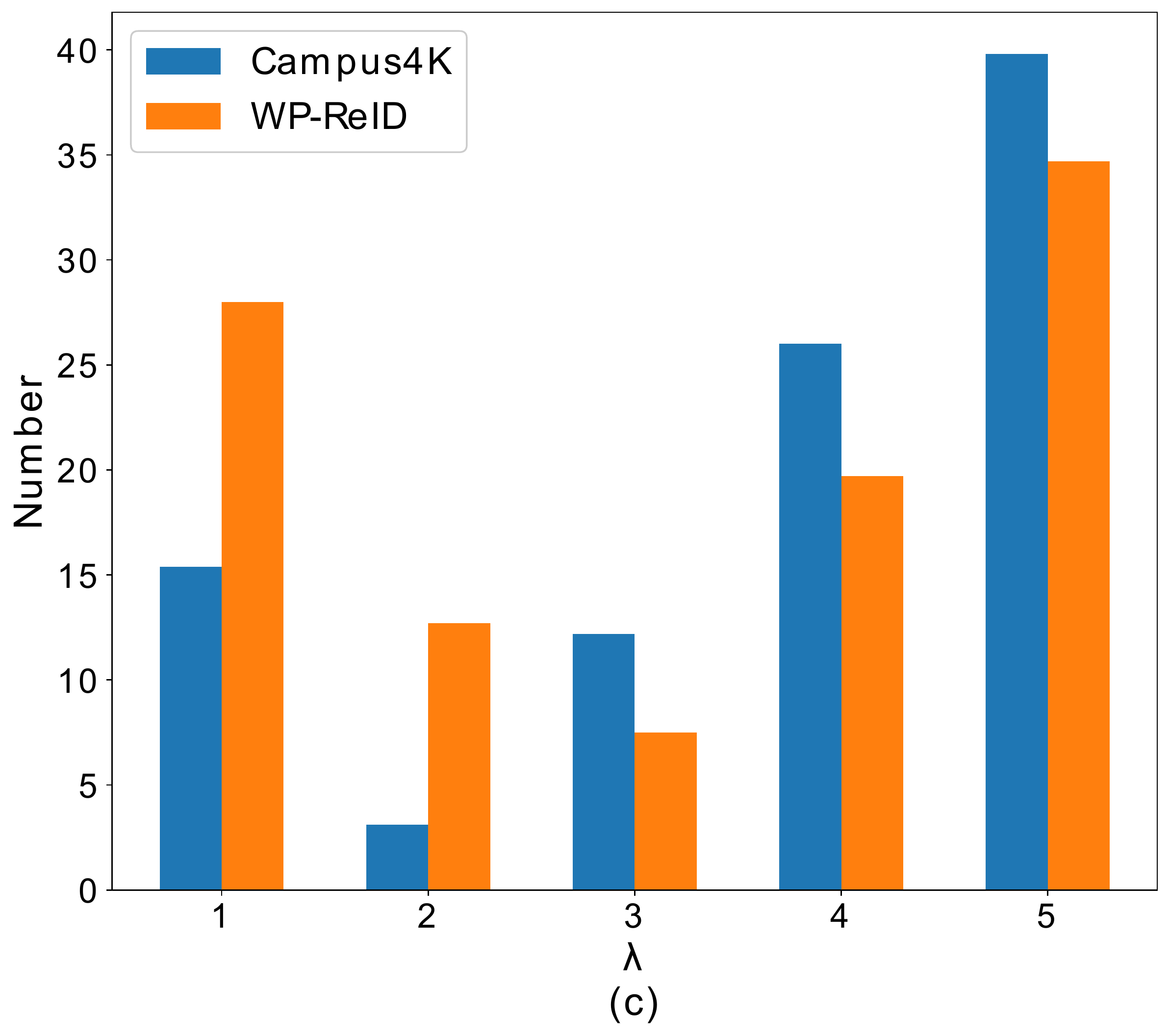}
    \end{center}
    \caption{(a) The accuracy of the clustering results and the accuracy of the multimodal data association between videos and wireless trajectories in MMDA. We use adjusted mutual information (AMI) as the evaluation metric of pseudo-label prediction and clustering.  
    (b) The influence of $\lambda$ in MMDA on the mAP accuracy of UMTF.  
     \emph{$K_m = K_m^{\mathrm{GT}}$} means that for $m^\mathrm{th}$ wireless trajectory, $K_m$ is set to the ground truth person number $K_m^{\mathrm{GT}}$ related with it, which is the upper bound of MMDA.
     \emph{$K_m = \sum_m{K_m^{\mathrm{GT}}} / M$} denotes that for all wireless trajectories, the number of clusters is the same and is set to the ground truth average person number related to wireless trajectories.
    (c) The effect of $\lambda$ on the difference between the predicted number of clustering centers and the true person numbers.}
    \label{fig:mmda:ablation}
    \end{figure*}

   In Table~\ref{tab:mmgn:loss}, we evaluate the impact of MMGN using different combinations of loss functions on the performance of UMTF. The performance is the highest when MMGN only uses cross-entropy loss. When the triplet loss \cite{hermans2017defense} is adopted, the performance of UMTF drops significantly. This is because MMGN has only a small number of parameters. This allows the input of MMGN in one iteration to be the video features of the entire dataset, instead of using a small batch of data like the convolutional network $F(\cdot)$. In our experiments, we find that constructing triplets based on video features of the entire dataset results in a large number of hard triplets, which makes the training of MMGN less stable.

   \textbf{Prediction of pseudo labels.} In Fig.~\ref{fig:all:ablation}(a), we show the percentage of the number of video sequences assigned pseudo visual labels or pseudo multimodal labels to the total number of video sequences. Most of the video sequences are assigned labels and participate in the training process of the model. 
   In Fig.~\ref{fig:all:ablation}(b), we show the change of the accuracy of pseudo-label predictions in UMTF with the increase of the number of training epochs.
   The baseline method only considers visual data and generates pseudo visual labels, whose prediction accuracies are also given in this figure. Adjusted mutual information (AMI) is adopted as the evaluation metric.
   It can be seen that the prediction accuracies of the two types of pseudo-labels in UMTF are higher than the pseudo-label prediction accuracy of the baseline method. This shows that our method UMTF effectively improves the accuracy of pseudo-label prediction.
   Meanwhile, when the number of training epochs increases, the prediction accuracies of the two types of pseudo-labels of UMTF increase first and then stabilize in the overall trend.
   The accuracies of the final label predictions of our method on both datasets are satisfactory.

   \textbf{Influence of the number of wireless trajectories.} In Fig.~\ref{fig:all:ablation}(c), we show the effect of the number of wireless signals on the model performance on WP-ReID dataset. The number of video sequences is fixed. When the number of wireless data is 50\% in Fig.~\ref{fig:all:ablation}(c), we only randomly select half of the wireless trajectories to participate in the training of UMTF.
   The performances of the model $F(\cdot)$ and MMGN are all improved when the number of wireless trajectories increased. This is because the increase in the number of wireless trajectories gives us more clues for multimodal pseudo-label prediction and further reduces the impact of visual noise.
   
   \textbf{Influence of the sensing radius.} We also show the influence of the sensing radius of wireless trajectories on the performance of UMTF in Fig.~\ref{fig:utfn:ablation}(a). When the sensing radius is small, the number of sensed wireless fragments is small. 
   When the radius is large, the expansion of the sensing area increases the number of sensed wireless fragments. 
   Affected by the size of the sensing area and wireless positioning accuracy, when two cameras are close, there will be some noisy (false-alarm) wireless fragments. 
   For a noisy fragment, the wireless fragment is sensed, but the pedestrian does not appear in the monitoring area of the camera. 
   In WP-ReID, the two closest cameras are only 50 meters apart. 
   It can be seen that the model performs the best when the radius is 50 meters. 
   When the sensing radius continues to increase, the performance of the UMTF is stable despite a lot of noisy fragments.
   The use of visual similarity during clustering and the soft association strategy in MMDA reduce the interference of noisy fragments and ensure the stability of MMDA.

   \textbf{Mutual promotion of dual models.}
   In Fig.~\ref{fig:utfn:ablation}(b), we show the accuracy changes of model $F(\cdot)$ and MMGN during the training of UMTF on WP-ReID dataset. 
   It can be seen that the mAP accuracy of MMGN is higher than $F(\cdot)$. The final mAP accuracy of the them is close. The rank-1 accuracy of MMGN is consistently lower than that of $F(\cdot)$.
   To verify whether the MMGN drags down the training of $F(\cdot)$, in Fig.~\ref{fig:utfn:ablation}(c), we evaluate the impact of the number of epochs MMGN participates in training on the performance of the model $F(\cdot)$. The experimental results show that the performance of $F(\cdot)$ is the highest when the MMGN is fully involved in 80 training epochs. This may be because although the rank-1 accuracy of MMGN is sometimes lower than $F(\cdot)$, it can assign correct pseudo multimodal labels to some videos that cannot be distinguished correctly by visual data.

   \textbf{Analysis of MMDA.}
   In the multimodal data association strategy (MMDA), we cluster the video sequences and associate them with wireless trajectories. As shown in Fig.~\ref{fig:mmda:ablation}(a), we show the accuracy of clustering and the accuracy of multimodal data association as the number of training epoch increases. It can be seen that the accuracy of clustering and the accuracy of data association achieve satisfactory results. Especially on the large Campus4K dataset, the accuracies of clustering and data association are more stable and higher.

   \textbf{Influence of $\lambda$.} As shown in Fig.~\ref{fig:mmda:ablation}(b), we evaluate the impact of $\lambda$ on UMTF. According to Eq.~\eqref{equ:cluster:k}, $\lambda$ controls the proportional relationship between the number of clusters $K_m$ and the estimated number of pedestrians related with $m^\mathrm{th}$ trajectory. 
   The results in Fig.~\ref{fig:mmda:ablation}(b) show that the performances of UMTF are not sensitive to $K_m$.
   By roughly adjusting $\lambda$, satisfactory results are obtained on both two datasets.
   When $\lambda$ is set to 3 and 2, UMTF achieves the highest performances on WP-ReID and Campus4K, respectively.
   In \emph{$K_m = K_m^{\mathrm{GT}}$}, $K_m$ is set to the ground truth person number $K_m^{\mathrm{GT}}$, which is the upper bound of the performance of our method.
   The results show that the performance of our adaptive estimation of $K_m$ is very close to the upper bound. 
   Taking the performance of UMTF on Campus4K dataset as an example, the performance of UMTF is only 0.37\% lower than the results achieved with ground truth person labels when $\lambda = 2$.  In \emph{$K_m = \sum_m{K_m^{\mathrm{GT}}} / M$}, when we set $K_m$ to the same value for all wireless trajectories, the performances drop on two datasets, which shows that it is necessary to adjust the number of clusters adaptively for the related video sequences of different wireless trajectories.
   Besides, in Fig.~\ref{fig:mmda:ablation}(c), we give the difference between the predicted number of clustering centers and the true number of pedestrians. The experimental results show that the deviation of the predicted value from the true value is very small when $\lambda$ is set to a suitable value.
   This indicates that Eq.~\eqref{equ:cluster:k} is a good way to estimate the appropriate number of clusters for different wireless trajectories, rather than trying without clues.

   \newcommand{\tabincell}[2]{\begin{tabular}{@{}#1@{}}#2\end{tabular}}  
   
   \begin{table*}
    \caption{Performance comparison with the state-of-the-art methods on WP-ReID. When the source dataset is selected, the corresponding method uses the labeled source dataset to pretrain the model. The full-scene labeling means whether to use the GPS coordinate labels of each location in the monitoring scenes of WP-ReID. UMTF$^{\mathrm{CCUP}}$ refers to using the contrastive learning method CCUP \cite{dai2021cluster} as the baseline method to replace the original baseline model in UMTF. The hyperparameters of different methods are all adjusted to the optimum.}
      \label{tab:wpreid:sta}
      \small
      \begin{center}
      \begin{tabular}{l|ccc|cccc}
      \hline
      \multirow{3}{*}{Method} & \multicolumn{7}{c}{WP-ReID} \\
      \cline{2-8}
      & \tabincell{c}{Source\\Dataset} & \tabincell{c}{Wireless\\Data} & \tabincell{c}{Full-scene\\Labeling} & mAP & Rank-1 & Rank-5 & Rank-10\\
      \hline
      TKP \cite{gu2019temporal}                       & \checkmark & $\times$ & $\times$ & 26.1 & 57.2 & 73.6 & 80.1 \\
      TKP+RCPM \cite{liu2020vision}                   & \checkmark & \checkmark & \checkmark & 47.5 & 67.2  & 76.6 & 83.6\\
      STMP \cite{liu2019spatial}                      & \checkmark & $\times$ & $\times$ & 36.8 & 64.2  & 80.6 & 86.1\\
      STMP+RCPM \cite{liu2020vision}                  & \checkmark & \checkmark & \checkmark & 60.4 & 78.1  & 80.6 & 83.5\\
      SSG \cite{fu2019self}                           & \checkmark & $\times$ & $\times$ & 28.2 & 63.7  & 77.1 & 82.1\\
      SSG+RCPM \cite{liu2020vision}                   & \checkmark & \checkmark & \checkmark & 47.8 & 72.1 & 79.6 & 85.1  \\
      MMT \cite{ge2020mutual}                         & \checkmark & $\times$ & $\times$ & 39.1 & 72.6 & 82.1 & 86.1\\
      MMT+RCPM \cite{liu2020vision}                   & \checkmark & \checkmark & \checkmark & 61.6 & 77.1 & 82.1 & 86.6 \\
      \hline
      SpCL\cite{ge2020self}                           & $\times$ & $\times$ & $\times$ & 30.4 & 50.2 & 74.6 & 83.1 \\
      IICS\cite{xuan2021intra}                        & $\times$ & $\times$ & $\times$ & 25.5 & 46.3 & 75.1 & 83.6\\
      MCDSCE\cite{yang2021joint}                      & $\times$ & $\times$ & $\times$ & 27.3 & 51.2 & 67.2 & 72.6 \\
      ICE\cite{chen2021ice}                           & $\times$ & $\times$ & $\times$ & 43.8 & 68.2 & 80.6 & 86.6 \\
      CCUP\cite{dai2021cluster}                       & $\times$ & $\times$ & $\times$ & 35.7 & 63.7 & 77.6 & 81.1 \\
      \hline
      Baseline                                        & $\times$ & $\times$ & $\times$ & 54.7 & 85.6 & 94.0 & 97.0  \\
      UMTF                                            & $\times$ & \checkmark & $\times$& 66.3 & \textbf{91.0} & \textbf{96.0} & \textbf{97.0}  \\
      UMTF$^{\mathrm{CCUP}}$ & $\times$ & \checkmark & $\times$ & \textbf{72.2} & 84.6 & 91.5 & 94.5 \\
      \hline
      \end{tabular}
      \end{center}
      
      \end{table*}

      \begin{table}
        \caption{Performance comparison with the state-of-the-art unsupervised methods on Campus4K. UMTF$^{\mathrm{CCUP}}$ refers to using the contrastive learning method CCUP \cite{dai2021cluster} as the baseline method to replace the original baseline model in UMTF.}
         \label{tab:duke:sta}
         \small
         \begin{center}
         \begin{tabular}{l|c|ccc}
         \hline
         \multirow{2}{*}{Method} & \multirow{2}{*}{Reference} & \multicolumn{3}{c}{Campus4K} \\
         \cline{3-5}
         & & mAP & R1 & R5  \\
         \hline
         TASTR\cite{xie2020progressive} & TMM'20 & 84.4 & 82.7 & 94.6 \\
         SpCL\cite{ge2020self} & NeurIPS'20 & 54.8 & 48.8 & 77.6 \\
         ICE\cite{chen2021ice} & ICCV'21 &  67.1 & 64.4 & 84.5 \\
         IICS\cite{xuan2021intra} & CVPR'21 & 67.0 & 64.5 & 85.7 \\
         MCDSCE\cite{yang2021joint} & CVPR'21 & 77.7 & 74.5 & 91.9 \\
         CCUP\cite{dai2021cluster} & arXiv'21 & 88.3 & 85.8 & 97.3 \\
         \hline
         Baseline & ours & 80.8 & 78.8 & 93.7 \\
         UMTF & ours & 86.6 & 85.4 & 95.5 \\
         UMTF$^{\mathrm{CCUP}}$ & ours & \textbf{93.7} & \textbf{92.9} & \textbf{98.8} \\
         \hline
         \end{tabular}
         \end{center}
         \end{table}

   \subsection{Comparison to the State-of-the-art Methods}
   
   We compare the performance of our approach UMTF with other state-of-the-art methods on WP-ReID dataset in Table~\ref{tab:wpreid:sta}. 
   The method RCPM \cite{liu2020vision} uses the GPS labels of each location of each scene to obtain the coordinate trajectory of the video sequence and calculate the distance to the wireless trajectory.
   Unlike \cite{liu2020vision}, our setting only needs to know the positions of cameras.
   The experimental results show that even if our model is not pre-trained with the labeled source dataset and does not use exhaustive scene labeling, our method still obtains the higher performances. 
   This shows that our setting for multimodal person re-identification is very effective. 

   As shown in Table~\ref{tab:duke:sta}, compared with the existing unsupervised methods that only rely on visual data, our method has obvious performance advantages on Campus4K dataset. This means that the introduction of wireless trajectories to assist unsupervised person re-identification is useful and necessary. In other words, wireless data and visual data can complement each other to reduce the influence of various interference factors.

   To further verify the generality of our method, we present the performance of our method UMTF$^{\mathrm{CCUP}}$ that uses the contrastive learning method CCUP \cite{dai2021cluster} as the baseline method. In UMTF$^{\mathrm{CCUP}}$, the loss functions and the training setting of model $F(\cdot)$ in the original UMTF are replaced by that of the method CCUP \cite{dai2021cluster}. 
   We also present the performance of existing unsupervised clustering-based method \cite{xie2020progressive,yang2021joint,xuan2021intra} and contrastive learning methods \cite{ge2020self,chen2021ice,dai2021cluster}. 
   Compared to the original baseline method, the new baseline CCUP performs better on Campus4K but worse on WP-ReID. This is because the number of videos of each pedestrian on WP-ReID is quite different. The imbalance of data brings great challenges to the contrastive learning method.

   It can be seen that our method UMTF$^{\mathrm{CCUP}}$ achieves the highest performance compared to these methods. For UMTF$^{\mathrm{CCUP}}$, compared with the new baseline CCUP, the mAP accuracy is improved by 36.5\%. This is due to the addition of MMDA and MMGN, which enable wireless data to assist the training of visual models and reduce the impact of visual noise. This also shows that our unsupervised multimodal training framework (UMTF) is general and still effective when replacing baseline visual method. 
   
   \section{Conclusion}\label{sec:conclusion}
   In this paper, we propose to use wireless trajectories to assist unsupervised person re-identification under weak scene labeling, \emph{i.e.}, we only need to know the locations of the cameras without labeling the latitude and longitude of each location in the monitoring scenes of cameras. We devise a new unsupervised multimodal training framework (UMTF) to train networks with both wireless data and visual data. UMTF mainly contains a multimodal data association strategy (MMDA) and a multimodal graph neural network (MMGN).
   MMDA calculates the wireless similarity measurement between video sequences by integrating the multimodal data. MMGN uses the specially designed multimodal graph convolutional module (MGM) to pass messages in the video graph. Through the collaboration between the various modules, UMTF obtains a better person representation model for re-identification. Extensive experiments on WP-ReID and Campus4K prove the effectiveness of UMTF.

   \section*{Acknowledgments}
   This work was supported in part by the National Natural Science Foundation of China under Contract U20A20183, 61836011 and 62021001, and in part by the Youth Innovation Promotion Association CAS under Grant 2018497. It was also supported by the GPU cluster built by MCC Lab of Information Science and Technology Institution, USTC.

\ifCLASSOPTIONcaptionsoff
  \newpage
\fi

\bibliographystyle{IEEEtran}
\bibliography{bare_jrnl_compsoc}

\begin{IEEEbiography}[{\includegraphics[width=1in,height=1.25in,clip,keepaspectratio]{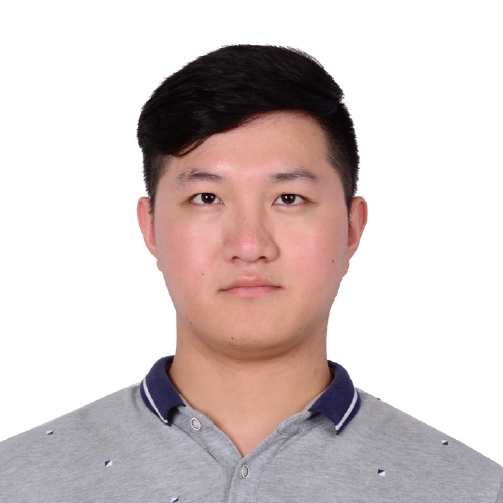}}]{Yiheng Liu}
  received the B.E. and Ph.D. degrees in electronic information engineering from the University of Science and Technology of China (USTC), Hefei, China, in 2017 and 2022.
  
  His research interests include person re-identification and computer vision.
  \end{IEEEbiography}
  
  \begin{IEEEbiography}[{\includegraphics[width=1in,height=1.25in,clip,keepaspectratio]{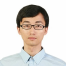}}]{Wengang Zhou}
    (S'20) received the B.E. degree in electronic information engineering from Wuhan University, China, in 2006, and the Ph.D. degree in electronic engineering and information science from University of Science and Technology of China (USTC), China, in 2011. From September 2011 to September 2013, he worked as a postdoc researcher in Computer Science Department at the University of Texas at San Antonio. He is currently a Professor at the EEIS Department, USTC. His research interests include multimedia information retrieval, computer vision, and computer game. In those fields, he has published over 100 papers in IEEE/ACM Transactions and CCF Tier-A International Conferences. He is the recepient of the Best Paper Award for ICIMCS 2012. He received the award for the Excellent Ph.D Supervisor of Chinese Society of Image and Graphics (CSIG) in 2021, and the award for the Excellent Ph.D Supervisor of Chinese Academy of Sciences (CAS) in 2022. He won the First Class Wu-Wenjun Award for Progress in Artificial Intelligence Technology in 2021. He served as the publication chair of IEEE ICME 2021 and won 2021 ICME Outstanding Service Award. He is currently an Associate Editor of IEEE Transactions on Multimedia. 

  \end{IEEEbiography}
  
  \begin{IEEEbiography}[{\includegraphics[width=1in,height=1.25in,clip,keepaspectratio]{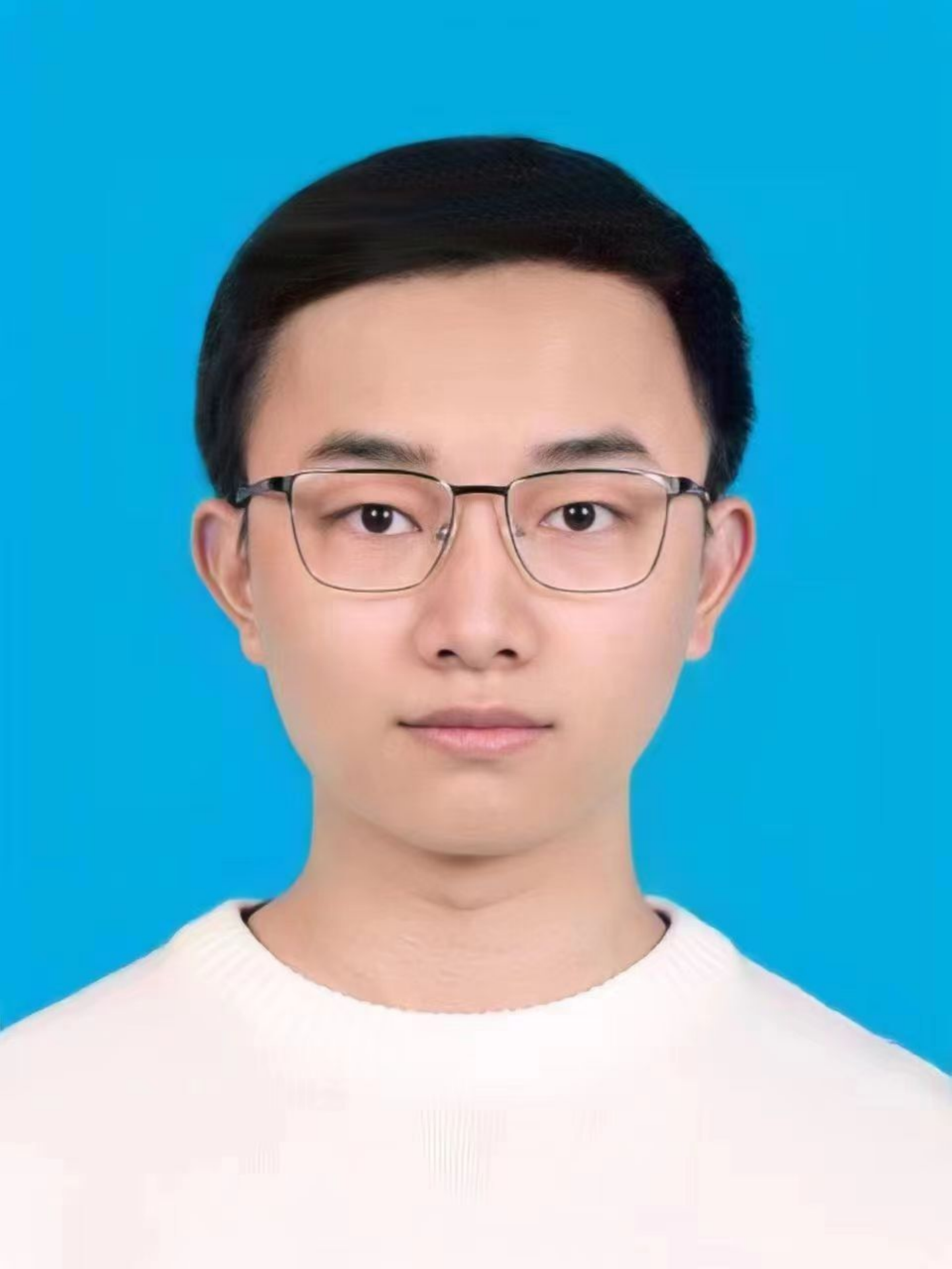}}]{Qiaokang Xie}
    received the B.E. and Ph.D. degrees in electronic information engineering from the University of Science and Technology of China (USTC), Hefei, China, in 2017 and 2022.
    
    His research interests include person re-identification and computer vision. 

  \end{IEEEbiography}
  
  \begin{IEEEbiography}[{\includegraphics[width=1in,height=1.25in,clip,keepaspectratio]{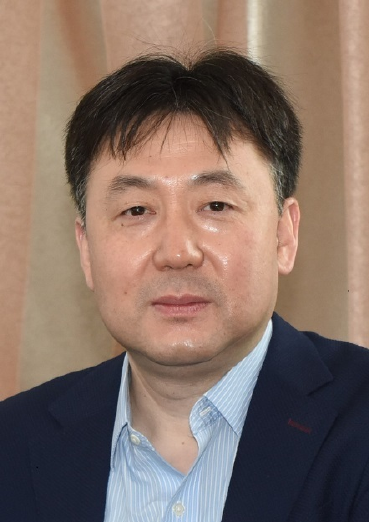}}]{Houqiang Li}
  (S'12, F'21) received the B.S., M.Eng., and Ph.D. degrees in electronic engineering from the University of Science and Technology of China, Hefei, China, in 1992, 1997, and 2000, respectively, where he is currently a Professor with the Department of Electronic Engineering and Information Science. 

  His research interests include image/video coding, image/video analysis, computer vision, reinforcement learning, etc.. He has authored and co-authored over 200 papers in journals and conferences. He is the winner of National Science Funds (NSFC) for Distinguished Young Scientists, the Distinguished Professor of Changjiang Scholars Program of China, and the Leading Scientist of Ten Thousand Talent Program of China. He is the associate editor (AE) of IEEE TMM, and served as the AE of IEEE TCSVT from 2010 to 2013. He served as the General Co-Chair of ICME 2021 and the TPC Co-Chair of VCIP 2010. He received the second class award of China National Award for Technological Invention in 2019, the second class award of China National Award for Natural Sciences in 2015, and the first class prize of Science and Technology Award of Anhui Province in 2012. He received the award for the Excellent Ph.D Supervisor of Chinese Academy of Sciences (CAS) for four times from 2013 to 2016. He was the recipient of the Best Paper Award for VCIP 2012, the recipient of the Best Paper Award for ICIMCS 2012, and the recipient of the Best Paper Award for ACM MUM in 2011.
  
  \end{IEEEbiography}

\end{document}